\pdfoutput=1
\documentclass{article}

\usepackage[main, final]{neurips_2026}

\usepackage[utf8]{inputenc}
\usepackage[T1]{fontenc}
\usepackage{hyperref}
\usepackage{url}
\usepackage{booktabs}
\usepackage{amsfonts}
\usepackage{nicefrac}
\usepackage{microtype}
\usepackage{xcolor}

\usepackage{graphicx}
\usepackage{subfigure}
\usepackage{algorithm}
\usepackage{algorithmic}
\usepackage{comment}
\usepackage{wrapfig}
\usepackage{caption}

\usepackage{amsmath}
\usepackage{amssymb}
\usepackage{mathtools}
\usepackage{amsthm}

\newcommand{\He}{{\cal H}_{\eta}}
\newcommand{\barHe}{\bar{{\cal H}}_{\eta}}

\usepackage{thmtools,thm-restate}

\usepackage[capitalize,noabbrev]{cleveref}

\theoremstyle{plain}
\newtheorem{theorem}{Theorem}[section]

\theoremstyle{definition}

\theoremstyle{remark}
\newtheorem{remark}[theorem]{Remark}

\usepackage[textsize=tiny]{todonotes}

\newcommand{\revise}[1]{{\color{black}#1}}

\title{Breaking the $\sqrt{d}$ Communication Barrier in Federated Sampling with Adaptive Hamiltonian Monte Carlo}

\author{%
  Jiajun Liang \\
  Independent Researcher\\
  \texttt{liangjiajun0529@gmail.com}
 \And
  Linxuan Wang \\
  Purdue University \\
   \texttt{wang6559@purdue.edu} \\
 \AND
 Guang Lin\\
 Purdue University \\
\texttt{guanglin@purdue.edu} \\
 \And
Qifan Song \\
Purdue University\\
 \texttt{qfsong@purdue.edu} \\
}

\begin{document}

\maketitle
\begin{abstract}
This work proposes the Adaptive Federated Hamiltonian Monte Carlo (AFHMC) for Bayesian federated learning (FL) tasks. The prior work on HMC applications for FL achieves $O(\sqrt{d}/\epsilon)$ communication cost under log-convex distributions, already demonstrating its advantage over the Langevin dynamic-based counterpart. Based on a refined analysis, this paper further reveals the benefit of HMC for the FL regime by incorporating a control mechanism for local node shifts. We prove that AFHMC at most requires $O(\{d/\epsilon^2\}^{1/3})$ communication cost up to a logarithmic term, which is significantly better than the existing results. Moreover, the proposed method adapts to client heterogeneity. Under low-heterogeneity settings (i.e., the target sampling precision $\epsilon^2$ is higher than the heterogeneity level), the communication rate of AFHMC can be as good as $O(\log(d/\epsilon^2))$, resembling the communication costs for FL optimization tools. To the best of our knowledge, this provides the first logarithmic communication complexity result for federated sampling, \revise{even though it requires low heterogeneity.
Numerically, we also justify AFHMC's advantage under the low-heterogeneity regime.}
Our code is available at \revise{{\small\url{https://github.com/laaaarrywang/Adaptive-Federated-HMC}}}.

\end{abstract}

\section{Introduction}

Federated learning (FL) provides a collaborative training paradigm in which multiple clients jointly learn a global model without sharing their raw data, thereby preserving data privacy while leveraging distributed data resources \citep{mmr+17,bonawitz2019towards,li2020federated}.
{A typical FL algorithm involves iterative rounds of local training followed by aggregating updates via node-to-center communications. When FL is applied to real-time systems, e.g., low-earth orbit (LEO) satellite constellations \citep{chen2022satellite}, it is critical that the FL algorithm design aligns well with the system constraints, such as limited connection bandwidth/windows and high communication latency. Therefore, a successful FL framework must be communication-efficient and tolerate low aggregation frequency.}
Despite its empirical success in large-scale applications, most existing FL algorithms \cite[e.g.,][]{konevcny2015federated,konevcny2016federated,LS20,wang2020federated}, are designed within a frequentist framework and focus on point estimation, which makes it difficult to quantify uncertainty in model predictions and parameters. Bayesian federated learning (BFL) addresses this limitation by inferring the full posterior, thus enabling principled uncertainty quantification and robust decision-making under data heterogeneity and limited local samples. However, scalable Bayesian FL inference remains challenging due to multiple factors: (i) high-dimensional parameter spaces, (ii) heterogeneous and potentially non-IID client data, and (iii) strict communication and privacy constraints.

Hamiltonian Monte Carlo (HMC, \cite{Duane87,Neal12}) is a MCMC method for sampling from complex posteriors, which exploits gradient information and an auxiliary momentum variable to construct proposals that can efficiently explore high-dimensional distributions. Compared with first-order methods such as stochastic gradient Langevin dynamics (SGLD, \cite{Welling11,sato2014approximation}), HMC and its stochastic variants often achieve superior mixing and lower asymptotic variance
but its second-order nature also makes theoretical analysis and distributed implementation more involved.
Extending HMC to the federated setting is nontrivial, because naive combinations of stochastic HMC with FedAvg-style aggregation \citep{mmr+17} can easily suffer from bias accumulation, instability of global dynamics, and inefficient communication.

To address these challenges, recent work proposes the Federated Averaging stochastic Hamiltonian Monte Carlo (FA-HMC, \cite{liang2025bayesian}) algorithm, which integrates stochastic HMC updates on local clients with periodic aggregation of both positions and momenta at a central server.
\cite{liang2025bayesian} showed that FA-HMC requires  $O(\sqrt{d}/\epsilon)$ communication cost (equivalently, $O(\sqrt{d}/\epsilon)$ updates occur in the center node) to achieve $\epsilon$ \revise{Wasserstein} sampling error. In contrast, under non-distributed settings, HMC achieves a tight $O(\{d/\epsilon^2\}^{1/4})$ gradient computation cost \citep{Mangoubi18_leapfrog}.
It naturally leads to the question: can we further close the gap between the communication cost in FL paradigm and the computation cost in the non-FL paradigm, in terms of the dimension dependency?

This work introduces an adaptive variant, Adaptive Federated Hamiltonian Monte Carlo (AFHMC), which incorporates surrogate leapfrog steps and an additional global shift vector to better align local and global dynamics. We conduct refine analysis which explicitly characterizes the discrepancy between surrogate and exact leapfrog dynamics, and establish non-asymptotic convergence guarantees under standard smoothness and strong convexity assumptions.
AFHMC can achieve improved dependence on the number of clients, communication rounds, and stochastic gradient noise level, and our experiments demonstrate competitive or superior performance compared with existing Bayesian federated baselines.

\section{Related Works and Our Contributions}

\paragraph{Federated optimization}
Early studies, such as FedAvg~\citep{mmr+17}, established convergence guarantees for smooth and strongly convex objectives, typically achieving sublinear rates in the presence of data heterogeneity. To address this challenge, subsequent work introduced variance reduction and control variate techniques, such as SCAFFOLD~\citep{kkm+20} and its variants, which improve convergence by mitigating client drift and reducing bias induced by heterogeneous local updates. In particular, \cite{mitra2021linear} provided the first result establishing linear convergence rates in terms of communication complexity using control variates, while \cite{pmlr-v162-mishchenko22b} further advanced this line of research by proposing randomized algorithms that achieve constant communication complexity under suitable conditions.

\paragraph{Bayesian federated learning.} Bayesian extensions of federated learning aim to characterize posterior uncertainty over model parameters or predictions while respecting client-level privacy and communication constraints. Existing approaches include variational Bayes formulations \citep[e.g.,][]{zhang2022personalized,guo2023federated}, hierarchical Bayesian models for heterogeneous clients \citep[e.g.,][]{vedadi2024federated, kim2025fedhb}, and Bayesian variants of federated averaging where each client maintains local posterior approximations that are periodically aggregated \citep[e.g.,][]{al-shedivat2021federated,chen2021fedbe,LiuBFL2024}. These methods often rely on factorized approximations or conjugate priors to keep the communication and computation overhead manageable.%
Our work instead targets exact (or asymptotically exact) MCMC-based Bayesian inference in federated environments, and develops HMC-based algorithms that can operate under realistic communication budgets.

\paragraph{HMC and stochastic variants.} HMC is an efficient sampler for complex high-dimensional Bayesian inference, with rigorous guarantees on convergence and mixing under smoothness and curvature conditions of the target distribution \citep{Irreducibility_HMC,mangoubi2017rapid, mangoubi2021mixing}. To scale HMC to large datasets, stochastic gradient HMC (SGHMC \cite{chen2014stochastic}) utilizes mini-batch estimates, typically introducing extra friction or correction terms to control the additional stochastic noise.
Subsequent works have investigated the convergence behavior of SGHMC \citep{zou2021convergence}, variance-reduction techniques \citep{zou2019stochastic}, low-precision sampling techniques \citep{wang2024enhancing} and etc.
These developments, however, are primarily designed for centralized settings, and their direct application to federated learning is hindered by client heterogeneity and limited access to the full dataset.

\paragraph{Adaptive and geometry-aware HMC.} A line of research has explored adaptive HMC schemes that adjust hyperparameters such as step sizes, mass matrices, and integrator structures using information gathered during sampling. Examples include \cite{no_U_turn_sampler,hoffman2021adaptive,hirt2021entropy,fernandez2016adaptive, girolami2011riemann},
all aiming to enhance stability and sampling efficiency in complex geometries. More recently, adaptive tuning strategies have been proposed to automatically select splitting integrators and hyperparameter ranges for (generalized) HMC, reducing reliance on heuristic tuning in practice \citep{Akhmatskaya2026Adaptive,nagar2024adaptive}. Our AFHMC algorithm is conceptually related to these works in that it uses a surrogate Hamiltonian flow informed by global information, but it is tailored to the federated setting, where the adaptation must be communication-efficient and robust to client heterogeneity.

Beyond existing results, the main contributions of this paper are summarized as follows:

1. We propose AFHMC, an adaptive HMC algorithm for FL tasks based on surrogate leapfrog updates. The new algorithm introduces a global drift control variate to address the heterogeneity among local gradient update trajectories.

2.  We introduce a principled framework to analyze divergence errors induced by approximate Hamiltonian dynamics.

3. Under strong convexity, we derive non-asymptotic error bounds that quantify the interplay between the surrogate leapfrog integrator and data heterogeneity. In particular, we prove that AFHMC achieves $O(\{d/\epsilon^2\}^{1/3})$ communication cost under extreme heterogeneity, and $O(\log(d/\epsilon^2))$ communication cost under low heterogeneity. To the best of our knowledge, both rates of communication complexity have never been achieved by prior federated sampling algorithms. Not only do we obtain a \revise{ better} communication rate, but we also don't rely on stringent Hessian smoothness conditions that are commonly required to achieve an advanced rate \citep[e.g.,][]{liang2025bayesian, Mangoubi18_leapfrog,plassier2023federated}. \revise{ Please refer to Table \ref{tab:compare} for a comprehensive comparison against prior theoretical works. Note that if we indeed enforce an extra Hessian smoothness condition, we could technically obtain even better results.}

4. We empirically demonstrate that the proposed method achieves improved computational efficiency and stability in distributed and approximate-gradient settings.

\section{Motivation and Algorithm}

\begin{figure}[t]
\captionsetup[figure]{skip=0pt}
  \centering
  \includegraphics[width=1\linewidth]{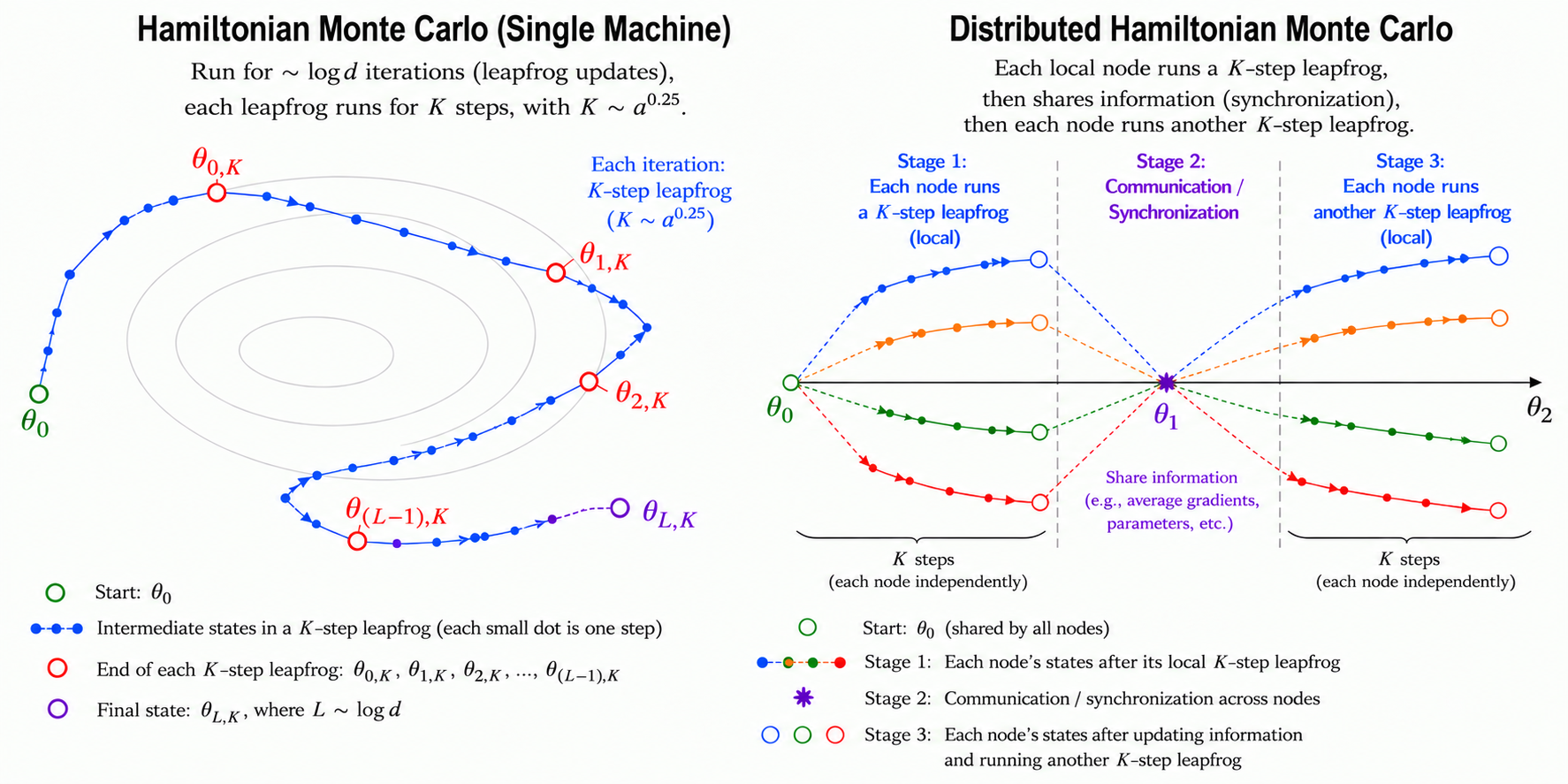}
  \caption{An illustration of the centralized HMC algorithm (left panel) and the federated HMC algorithm (right panel).}
  \label{fig:motivation}
\end{figure}

\revise{The goal of a standard sampling problem is to generate approximate samples from the target
posterior $\pi (\theta)\propto \exp(-f(\theta))$ where $f$ is the potential function. In the standard leapfrog HMC, as illustrated in Figure~\ref{fig:motivation}(left), the parameter $\theta$ } is evolved using a randomly sampled Gaussian momentum for $K$ leapfrog steps (defined later in \eqref{eq:lf}), after which the momentum is resampled before starting a new sequence of leapfrog updates. In \cite{Mangoubi18_leapfrog},  it is shown that performing $O(\log d)$  such leapfrog trajectories, each consisting of $O(\{d/\epsilon^2\}^{1/4})$ steps, suffices to ensure convergence of the task distribution. Therefore, the total gradient computational cost of the overall procedure is dominated by the number of leapfrog steps, leading to a total complexity of order $O(d^{1/4})$ up to logarithmic factors.

\revise{Let \([N]=\{1,\ldots,N\}\) index the clients. In federated settings, each client $c$ has a
local potential $f^{(c)}(\theta)$ with weight $w_c\ge0$ and thus global potential $f(\theta)=\sum_{c=1}^N w_c f^{(c)}(\theta)$,
where \(\sum_{c=1}^N w_c=1\).} To achieve logarithmic communication complexity in federated sampling, the above theoretical insight motivates us to integrate communication steps into the alternating leapfrog procedure, expecting that the order of FL communication cost matches $O(\log d)$ order of leapfrog trajectories in centralized HMC. To be concrete, as illustrated in Figure~\ref{fig:motivation} (right), we aim that each local node performs $K$ leapfrog steps followed by a communication round (e.g., averaging the local parameters across nodes), before proceeding to the next leapfrog phase. This corresponds to the FA-HMC algorithm \citep{liang2025bayesian} with $T=1$. Unfortunately, \cite{liang2025bayesian} establishes a lower bound showing that, under high federated heterogeneity, the communication cost of such an HMC scheme is at least on the order of $O(\{d/\epsilon^2\}^{1/2})$. Thus, additional non-trivial FL training designs are necessary to reach a better rate.

In any FL optimization tasks, a key challenge to achieve efficient communication is to control the local shift caused by heterogeneity across local nodes. Local gradient updates move the local parameter toward the location optimum, resulting in a global bias after aggregating local solutions. A breakthrough was introduced by \cite{kkm+20}, which introduced a control shift technique that reduces the communication cost of FedAvg  \citep{mmr+17} from $O(d/\epsilon^2)$ to $O(\log(d/\epsilon^2))$.
In the FL sampling regime, FA-LD \citep{Wei2024On}, a Langevin-counterpart of FedAvg also incurs a communication cost of $O(d/\epsilon^2)$. \cite{plassier2023federated} further improved the sampling algorithm by adapting the control shift technique of \citep{kkm+20}.

Inspired by existing works on local shift control, we propose an adaptive HMC federated update rule by introducing a drift variate. Given an energy function $F$, the HMC dynamics follow the Hamiltonian equations:
\begin{equation}\label{eq:hmcdynamic0}
       d\theta(t)=p(t)dt,\qquad dp(t)=-\nabla_\theta F(\theta(t))dt.
\end{equation}
The above continuous-time dynamic cannot be solved analytically. Commonly, one uses the leapfrog integrator to discretize the dynamics over a step size $\eta$ by
\begin{equation}\label{eq:lf}
\theta_{k+1}=\theta_k+\eta p_k-\frac{\eta^2}{2}\nabla F(\theta_k),\qquad p_{k+1}=p_k-\frac{\eta}{2} \nabla F(\theta_k)-\frac{\eta}{2}\nabla F(\theta_{k+1}).
\end{equation}
Under the FL setting, update algorithm (2) is running on local nodes, and $F$ represents a local energy function which can differ substantially from the global energy function $f$ due to node heterogeneity. To account for this potential gap between local $F$ and global $f$, we inject a bias correction term to the gradient and replace all $\nabla F(\cdot)$ terms in (\ref{eq:lf}) with $\nabla F(\cdot)-\nabla F(\theta_0)+\nabla f(\theta_0)$ (see, Algorithm \ref{alg:surHMC}). This bias correction term $\nabla f(\theta_0)-\nabla F(\theta_0)$ addresses the discrepancy between local and global losses, and the drift variate $\nabla f(\theta_0)$ needs to be broadcast to all local nodes.
This novel local update rule is named the surrogate leapfrog update.

The AFHMC algorithm consists of the following iterative steps: 1. After synchronizing parameter $\theta_t$ across all nodes, each node $c$ shares the local $\nabla f^{(c)}(\theta_t)$ to obtain a global drift variate $\nabla f(\theta_t)$. 2. All local nodes run $K$ surrogate leapfrog steps, with the shared global drift variate, initialized from the same position and momentum. 3. All nodes share and synchronize the local parameter $\theta_{t+1}$, after $K$ leapfrog steps. The algorithm can be found in Algorithm \ref{alg:AFAHMC}.
In contrast to FA-HMC \citep{liang2025bayesian}, which runs multiple local HMC steps where each HMC step consists of $K$ leapfrog updates, the proposed AFHMC only runs one single local HMC step (i.e., $K$ updates of surrogate leapfrog, see Figure \ref{fig:motivation}). As discussed previously, this design is to ensure minimal communication cost, because every local HMC step will need a fresh new drift control variate, which requires communication.

The AFHMC algorithm can be easily implemented with stochastic gradients for acceleration purposes, similarly to \cite{liang2025bayesian}, but our analysis in this paper focuses on the use of the exact gradient.

\begin{algorithm}[htb]
   \caption{Surrogate leapfrog approximation $\widetilde{h}_{\rm LF}(\cdot)$ }
   \label{alg:surHMC}
\begin{algorithmic}
   \STATE {\bfseries Input:}  energy function $F(\cdot)$, initial position $\theta_0$, drift vector $\nabla f(\theta_0)$, momentum $p_0$, stepsize $\eta$, leapfrog step $K$
   \FOR{$k=0$ {\bfseries to} $K-1$}
  \STATE  $\theta_{k+1}=\theta_k+\eta p_k-\frac{\eta^2}{2}(\nabla F(\theta_k)-\nabla F (\theta_0)+\nabla f (\theta_0))$
   \STATE     $p_{k+1}=p_k-\frac{\eta}{2} \nabla F(\theta_k)-\frac{\eta}{2}\nabla F(\theta_{k+1})+\eta(\nabla F (\theta_0)-\nabla f (\theta_0))$
   \ENDFOR
    \STATE {\bfseries Output:} $\widetilde{h}_{\rm LF}(F,\theta_0,\nabla f(\theta_0),p_0,\eta,K)=\theta_K$
\end{algorithmic}
\end{algorithm}

\begin{algorithm}[htb]
	\caption{Adaptive federated hamiltonian monte carlo (AFHMC) algorithm}\label{alg:AFAHMC}
	\begin{algorithmic}
		\STATE {\bfseries Input:} initial position $\theta_0$,  stepsize $\eta$;  leapfrog update step $K$
  \STATE Each local node receives the initial parameter $\theta_0$ from the center
		\FOR{$t=0$ {\bfseries to} $t_{\max}$}
  \STATE {\bf $1st$ broadcast:} The $c$-th local node sends $\nabla f^{(c)}(\theta_t)$ to the center, for all $c=1,\dots,N$;
  \STATE {\bf $1st$ communicate:} The center node samples Gaussian momentum $p^t$, and sends $p_t$ and drift vector $\nabla f(\theta_t):=\sum_{c=1}^Nw_c\nabla f^{(c)}(\theta_t)$ to all local nodes;
  \FOR{$c=1,2 \ldots,N$}
   \STATE {\bf  Initialize} $\theta^{(c)}_{t,0}=\theta_t$, $p_{t,0}^{(c)}=p_t$;
   \STATE {\bf  update }    $\theta^{(c)}_{t,K}=\widetilde{h}_{\rm LF}(f^{(c)},\theta^{(c)}_{t,0},\nabla f(\theta_t),p_{t,0}^{(c)},\eta,K)$;
  \ENDFOR
 \STATE {\bf $2nd$ broadcast:} The $c$-th local node sends $\theta^{(c)}_{t,K}$ to the center for all $c=1,\dots,N$;
 \STATE {\bf $2nd$ communicate:} The center node sends back $\theta_{t+1}:=\sum_{c=1}^Nw_c\theta_{t,K}^{(c)}$ to all local nodes;
  \ENDFOR
   \STATE {\bfseries Output:} $\{\theta_t:t=0,\ldots, t_{\max}\}$
	\end{algorithmic}
\end{algorithm}

\section{Convergence Analysis}

Let $z^c_{i}, 1\leq i\leq n_c, 1\leq c\leq N$ be the local data at the $c$-th node, $\theta\in \mathbb{R}^d$ and $\ell(\theta; z^c_{i})$ be the negative log-likelihood (or loss) function. Define $w_c=n_c/\sum_{c=1}^N n_c$, and $f^{(c)}(\theta):=\sum_{i=1}^{n_c}\ell(\theta; z^c_{i})/w_c$ is the local loss function of accessible to the $c$-th local node. The goal is to simulate the global target distribution $\pi(\theta)\propto \exp(-{f(\theta)})$, where $f(\theta)=\sum_{c=1}^N w_cf^{(c)}(\theta)$, $w_c\geq 0$ with $\sum_c w_c = 1$.

This section will investigate the theoretical \revise{properties} of AFHMC algorithm. Before stating our main results, we specific the model assumption as follows:

\begin{restatable}[$m$-Strongly Convex]{assumption}{assumconvex}\label{assum:convex}
For each $c=1,2,\ldots, N$, $f^{(c)}$ is $m$-strongly convex for some $\mu>0$, i.e., $\forall x, y \in \mathbb{R}^{d}$
$
f^{(c)}(y)\geq f^{(c)}(x)+\langle\nabla f^{(c)}(x), y-x\rangle+\frac{m}{2}\|y-x\|_{2}^{2}.
$
\end{restatable}
\begin{restatable}[$M$-Smoothness]{assumption}{assumsmooth}\label{assum:smooth}
 For each $c=1,2,\ldots, N$, $f^{(c)}$ is $M$-smooth for some $M>0$, i.e., $\forall x, y \in \mathbb{R}^{d}$, we have
$
\|\nabla f^{(c)}(y)-\nabla f^{(c)}(x)\|\leq M\|x-y\|.%
$
\end{restatable}

The assumptions are fairly standard in the theoretical literature of gradient-based MCMC algorithms \citep[e.g.,][]{dk17,Mangoubi18_leapfrog,dk19,erdogdu2021convergence}. Unlike prior advanced results \citep[e.g.,][]{dk19,liang2025bayesian,plassier2023federated,Mangoubi18_leapfrog}, we don't require any second-order smoothness (e.g., Hessian) of the energy functions.

Our first result provides a non-asymptotic bound, studying how the gap between $\theta_t$ and $\theta_{t}^{\pi}$ contracts, where $\theta_{t}^{\pi}$ is a coupled stationary continuous HMC process satisfying $\theta^{\pi}_t\sim\pi$ for all $t\geq0$ (refer to Appendix Section \ref{sec:a} for formal definitions).
\begin{restatable}[Convergence]{theorem}{thmAFLHMC}\label{thm:AFLHMC}
Under Assumptions \ref{assum:convex}-\ref{assum:smooth},   there exist sufficient large constants $C_0,C_1,C_2$ such that $\{\theta_t\}_t$ satisfies that, for $K\eta\leq \sqrt{m}/(C_0M)\leq \sqrt{m}/(2M)$,
\[
    \mathbb{E} \|\theta_{t+1}-\theta^\pi_{t+1}\|^2
     \leq\left(1-\frac{m(K\eta)^2}{16}\right)\mathbb{E}\|\theta_t-\theta^\pi_t\|^2+C_1K^2\eta^4\frac{M^2}{m}d+C_2(K\eta)^8\frac{M^2}{m}H
\]
where $H:=\max_{i\leq K}\sum_{c=1}^Nw_c\mathbb{E}\|\frac{1}{i\eta}(\nabla f^{(c)}(\barHe^i(\theta^\pi))-\nabla f^{(c)}(\theta^\pi)+\nabla f(\theta^\pi)-\nabla f(\barHe^i(\theta^\pi))\|^2$ is a metric quantifying the federated
 heterogeneity.
\end{restatable}

The $\barHe^i(\theta)$ (formally defined in Appendix Section \ref{sec:a}) denotes the output by running $i\eta$-length ideal continuous HMC dynamic (i.e., (\ref{eq:hmcdynamic}) with $F$ being replaced by $f$) initialized from $\theta$.
\begin{remark}\label{rem:1}
{ $H$ is a metric only related to the target distribution and local distribution. It characterizes the client-level gradient heterogeneity under the target distribution.
$H$ is a theoretical quantity usually unknown in practice. But in terms of its dependency on the dimension, due to the nature of the squared $L_2$ norm, one can prove that $H = O(d)$. \revise{The proof of this bound and  examples of calculating the exact $H$ are given in Appendix Section~\ref{section:H}.}}
\end{remark}

\paragraph{Sketch of proof}
We decompose $\|\theta_{t+1}-\theta^\pi_{t+1}\|^2$ into two terms:
$\|\barHe^K(\theta_t)-\theta^\pi_{t+1}\|^2$ and $\| \theta_{t+1}-\barHe^K(\theta_t)\|^2$. The first term is subject to the classic contraction theorem of ideal HMC dynamics. The second term can be characterized by  (i) the divergence between $\theta_{t,k}^{(c)}$ and $\sum_{c=1}^Nw_c \theta^{(c)}_{t,k}$, and (ii) the divergence between ideal HMC dynamics $\barHe^i(\theta)$ and the discretized HMC dynamics $\He^i(\theta)$. Thanks to the local shift control mechanism, we establish a new technical bound (Lemma \ref{lem:diffB})) to study the surrogate error  and give
$\mathbb{E}\|\theta^{(c)}_{t,k}-\sum_{c=1}^Nw_c\theta_{t,k}\|^2\approx O((K\eta)^6H)$ , and bound the divergence caused by discretizations by $O(\eta^4d)$  (via Lemma \ref{lem:AveLocB}).

Based on the recursive contraction result in Theorem \ref{thm:AFLHMC}, one can easily derive the convergence rate for $\mathbb{E} \|\theta_{t+1}-\theta^\pi_{t+1}\|^2$. Let $\pi_t$ be the marginal distribution of $\theta_t$ and ${\cal W}_2(\mu,\nu)$ denote the Wassertein-2 distance between probability measures $\mu$ and $\nu$. Then, by definition, ${\cal W}_2^2(\pi_t,\pi)=\inf_{\theta\sim\pi}\mathbb{E}\|\theta_t-\theta\|^2\leq \mathbb{E} \|\theta_{t}-\theta^\pi_{t}\|^2$, leading to
\[
{\cal W}_2^2(\pi_{t},\pi)\leq \left(1-\frac{m(K\eta)^2}{16}\right)^t {\cal W}_2^2(\pi_{0},\pi)+C_1'\eta^2\frac{M^2d}{m^2}+C_2'K^6\eta^6\frac{M^2H}{m^2}.
\]
This leads to our next theorem, which studies the communication cost rate for AFHMC.
\begin{restatable}[]{theorem}{thmconvergratesg}\label{thm:convergrate:sg}
Under Assumptions~\ref{assum:convex}-\ref{assum:smooth} and if $H\gg M^4\epsilon^2/m>0$, then
there exists some small absolute constant $C$ such that if we choose%
\[
 K\eta = C\left(\frac{\epsilon^2m^2}{HM^2}\right)^{1/6}\ll C\frac{\sqrt m}{M}\quad\mbox{ and }\quad \eta = C\left(\frac{\epsilon^2m^2}{M^2d}\right)^{1/2},
\]
 ${\cal W}_2(\pi_{t_\epsilon},\pi)\leq \epsilon$ can be attained with communication times
\[
t_\epsilon\asymp  \left(\frac{HM^2}{\epsilon^2m^5}\right)^{1/3}\log( {\cal W}_2^2(\pi_{0},\pi)/\epsilon^2).
\] The corresponding local iteration number is
\[
K\asymp \frac{M^{2/3}d^{1/2}}{\epsilon^{2/3}m^{2/3}H^{1/6}}.
\]

Moreover, if $H \ll M^4\epsilon^2/m$ (i.e., heterogeneity across local nodes is very small), it only requires
\[
t_\epsilon\asymp\log( {\cal W}_2^2(\pi_{0},\pi)/\epsilon^2),
\]
achieving a linear rate with respect to the communication cost.
\end{restatable}

For a heterogeneous situation, the communication cost $t_\epsilon$ implicitly depends on the model dimension via the quantity $H$.
As discussed in Remark \ref{rem:1}, $H$ can scale linearly with $d$, leading to $t_\epsilon\asymp (d/\epsilon^2)^{1/3}\log(d/\epsilon^2)$. For a homogeneous situation where $H\approx0$, the communication cost reduces to $O(\log(d/\epsilon^2))$. To the best of our knowledge, both rates have never been achieved by prior Bayesian FL methods in the literature.  Moreover, our analysis do not require extra stringent Hessian smoothness conditions that are commonly required to achieve an advanced rate e.g., Assumption A3 in \cite{plassier2023federated}, Assumption 3.3 in \cite{liang2025bayesian}. \revise{We compare all these results in Table~\ref{tab:compare}.
\begin{table}[t]
\centering
\resizebox{\textwidth}{!}{
\begin{tabular}{|c|c|c|c|}
\toprule
\textbf{Algorithm}
& \textbf{Condition beyond }
& \textbf{Rate of communication} & \textbf{Rate of gradients } \\
&\textbf{Assumptions~\ref{assum:convex}-\ref{assum:smooth}}&&\textbf{(up to logarithmic terms)}\\
\midrule
Centralized HMC
& Hessian Lipschitz under infinity  & none
& $(d/\epsilon^{2})^{1/4}$  \\
~\cite{Mangoubi18_leapfrog}&norm (Assum. 1 in their paper)&&\\
\midrule
VR-FALD
& Hessian Lipschitz under L2
& $(d/\epsilon^{2})^{1/2}$  & $(d/\epsilon^{2})^{1/2}$  \\
~\cite{plassier2023federated}&norm (Assum. HX1 in their paper)&&\\
\midrule
FA-HMC
& Hessian Lipschitz under L2
& $(d/\epsilon^{2})^{1/2}$  & $(d/\epsilon^{2})^{1/2}$ \\
~\cite{liang2025bayesian}&norm (Assum. 3 in their paper)&&\\
\midrule
AFHMC (this paper)
& none
& $\log(d/\epsilon^{2})\sim(d/\epsilon^{2})^{1/3}$  & $(d/\epsilon^{2})^{1/2}\sim(d/\epsilon^{2})^{2/3}$ \\
\bottomrule
\end{tabular}
    }
    \caption{Comparison of Bayesian theoretical results: requires assumptions, rate of communication and rate of gradient computation to achieve $\epsilon$ Wasserstein convergence}\label{tab:compare}
\end{table}
}

This theorem can also be interpreted from a different perspective. AFHMC can have two training phases: in the first phase (fast convergence), the sampling error linearly decays until $ {\cal W}_2^2(\pi_{t},\pi)\sim mH/M^4$; in the second phase (slow convergence), it needs $O((d/\epsilon^2)^{1/3}\log(d/\epsilon^2))$ communications to reach $ {\cal W}_2^2(\pi_{t},\pi)=\epsilon^2=o(mH/M^4)$. Note that these two phases do require different tuning of $K$ and $\eta$, as shown by the proof of Theorem \ref{thm:convergrate:sg}.

\begin{remark}
  Our main theory is derived based on using the exact gradient updates. While it is of great interest to investigate the theoretical behavior when stochastic gradient is used instead, it poses great challenges to obtain a sharp rate that surpasses the Langevin counterpart. Such a result doesn't exist so far, even for the centralized HMC algorithm. With a lack of theoretical guarantee, we still believe in the advantage of AFHMC under noisy gradient, which is empirically justified by our numerical studies.
\end{remark}

\section{Numerical Evidences}\label{sec:numerical}
In this section, we present our numerical results, including both simulation studies and real data applications, to support our theoretical claims. We compare the empirical performance of AFHMC, FA-HMC and FA-LD, and demonstrate the superiority of the proposed AFHMC algorithm.
\revise{Throughout this section, a \emph{communication round} refers to one outer iteration of the federated algorithm. Each AFHMC round consists of two server exchanges (the gradient aggregation for the drift correction and the parameter aggregation after the surrogate trajectory; see the two broadcasts in Algorithm~\ref{alg:AFAHMC}), whereas each FA-HMC or FA-LD round consists of one. This constant factor does not change the communication complexity rates. For an empirical comparison at an equal number of server exchanges, we compare AFHMC at round $R$ with FA-HMC at round $2R$.}

\paragraph{Simulated Bayesian logistic regression.} Following \cite{liang2025bayesian,Mangoubi18_leapfrog}, we consider Bayesian logistic regression with a standard Gaussian prior on $\theta \in \mathbb{R}^d$. Given feature vectors $x_i\in\mathbb{R}^d$ and binary labels $y_i\in\{-1,+1\}$, the posterior is
\[
\pi(\theta) \propto \exp\left(-\frac{1}{2}\|\theta\|^2\right)
\prod_{i=1}^{n} \sigma(y_i x_i^\top \theta),
\]
where $\sigma(\cdot)$ denotes the sigmoid function. The $n=1000$ observations are split evenly across $N=20$ clients. We evaluate the samplers in two settings: (1) a high-dimensional case with $d=1000$ and exact local gradients; (2) a lower-dimensional case with $d=10$ with inaccurate gradients. Specifically, we add $\mathcal{N}(0,100I_d)$ noise to the exact gradients, simulating the stochastic mini-batching.

The algorithms (FA-LD, FA-HMC, and AFHMC) are implemented with the following tuning parameter settings.
FA-LD uses $T=10$ local Langevin steps (each Langevin step is equivalent to $K=1$ leapfrog step). FA-HMC uses $T=10$ local HMC steps, and each HMC step consists of $K_F$ classic leapfrog steps. We sweep $K_{\text{F}}$ over a grid centered around the trajectory-length rule $K_{\text{F}}\eta\approx \pi/3$ \citep{Mangoubi18_leapfrog}. AFHMC uses $K_{\text{A}}$ local surrogate leapfrog steps, and we consider two candidates of $K_{\text{A}}$:
(i) $K_{\mathrm{A}}=K_{\text{F}}$ such that AFHMC and FA-HMC match the HMC dynamic trajectory length, and (ii) $K_{\mathrm{A}}=10K_{\mathrm{F}}$ such that AFHMC and FA-HMC
match the same number of leapfrog steps between consecutive rounds of communication. For different $\eta$, the $K_{\mathrm{A}}$ and $K_{\text{F}}$ values that deliver the best performance are chosen. For each $(\text{algorithm},\eta,K)$ configuration, we run $M=20$ independent chains with a total leapfrog budget of $2\times 10^{6}$ and discard the first half of each chain as burn-in. We measure performance using the marginal error
\begin{equation}
  \mathrm{ME}
  \;=\; \frac{1}{d} \sum_{i=1}^{d}
        \mathcal{W}_1\!\left(\hat\mu_i,\, \mu_i^{\mathrm{MHMC}}\right),
  \label{eq:me}
\end{equation}
where $\hat\mu_i$ is the empirical marginal of the $i$-th coordinate produced by the candidate sampler, $\mu_i^{\mathrm{MHMC}}$ is the surrogate true posterior margins, approximated by a centralized Metropolis-adjusted HMC reference chain with $10^{7}$ steps, and $\mathcal{W}_1$ denotes the Wasserstein-1 distance. For each $(\text{algorithm},\eta)$ pair, we report the best-performing $K$ from the candidate grid; the full grid is provided in Appendix~\ref{app:fig1-grid}.
\begin{figure}[t]
  \centering
  \includegraphics[width=0.48\linewidth]{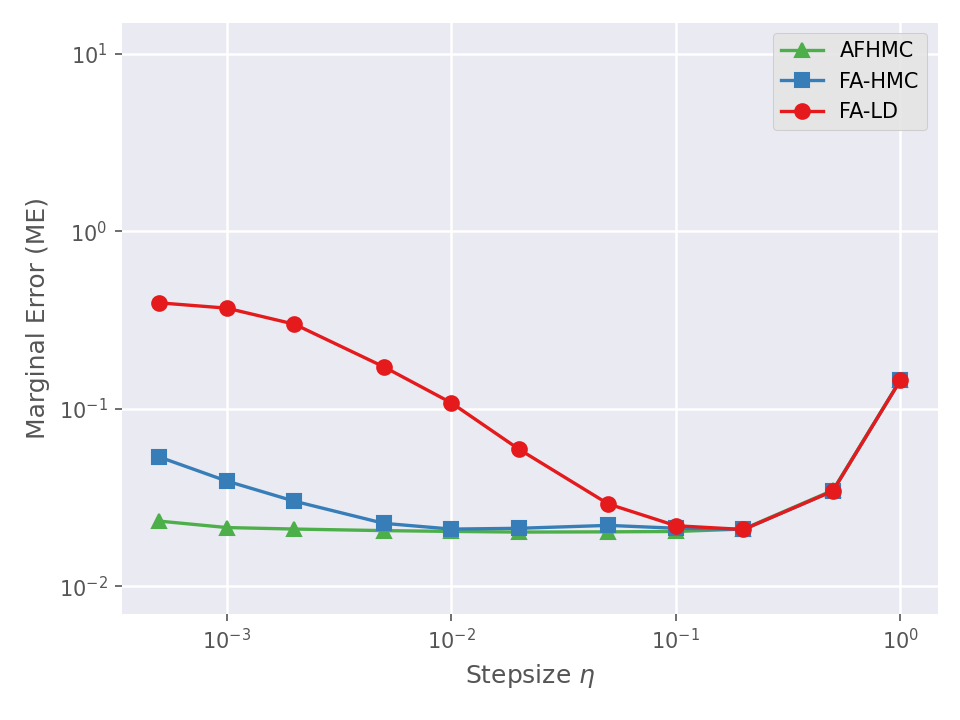}
  \hfill
  \includegraphics[width=0.48\linewidth]{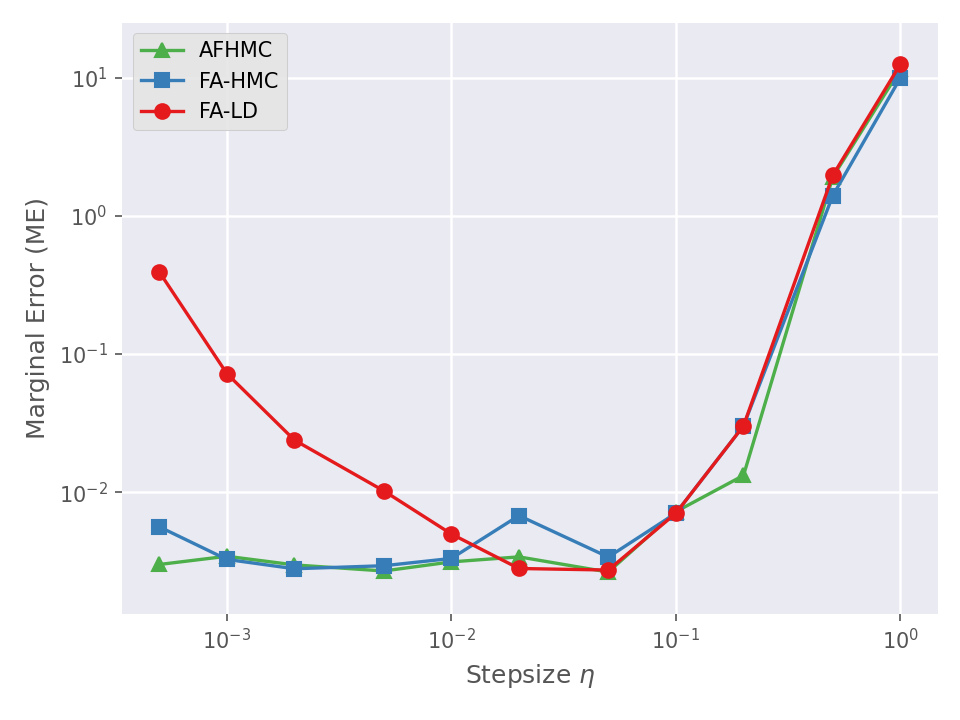}

  \caption{Simulated Bayesian logistic regression. Left: $d=1000$ with exact gradients. Right: $d=10$ with coordinate-wise $\mathcal{N}(0,100)$ gradient noise. Each panel reports marginal error relative to a centralized MHMC reference chain under a fixed total leapfrog budget. AFHMC is more stable across stepsizes and improves the small-stepsize behavior of FA-HMC and FA-LD.}
  \label{fig:logreg}
\end{figure}

Figure~\ref{fig:logreg} reports the resulting marginal error curves for the exact-gradient and noisy-gradient settings. In the exact-gradient high-dimensional setting, AFHMC maintains low marginal error over a broader range of small stepsizes: for example, at $\eta=5\times 10^{-4}$, AFHMC attains $\mathrm{ME}=2.3\times 10^{-2}$, compared with $5.3\times 10^{-2}$ for FA-HMC and $3.9\times 10^{-1}$ for FA-LD. In the noisy-gradient setting, AFHMC achieves the best overall marginal error in the sweep, $\mathrm{ME}=2.7\times 10^{-3}$, and remains competitive with the best tuned FA-HMC and FA-LD configurations at the other stepsizes.

\paragraph{Dimension scaling on heterogeneous Gaussian targets.}
Following the heterogeneous Gaussian experiment of
\cite{liang2025bayesian}, we study how the communication cost scales with dimension on a controlled federated Gaussian benchmark. There are $N=10$
clients. For half of the clients the local target is
$\mathcal{N}(20\mathbf{1}_d,I_d)$, and for the remaining clients it is
$\mathcal{N}(\mathbf{1}_d,4I_d)$. Averaging the local quadratic potentials gives
the global target $\mathcal{N}(16.2\mathbf{1}_d,1.6I_d)$. We vary
$d\in\{2,50,100,150,\ldots,500\}$ and measure convergence by the squared 2-Wasserstein distance to this global Gaussian target. Note that the empirical 2-Wasserstein distance is expensive to compute for high-dimensional data, therefore, we approximate the posterior distribution by fitting a multivariate Gaussian and compute the 2-Wasserstein distance between the fitted Gaussian distribution and the target distribution via the closed-form formula.

Suggested by the theoretical choice provided by Theorem \ref{thm:convergrate:sg}, for AFHMC, the leapfrog count and stepsize are scaled with dimension as
\[
  K(d)=\max\left\{1,\left\lfloor 50(d/100)^{1/3}\right\rceil\right\},
  \qquad
  \eta(d)=0.006(d/100)^{-1/2},
\]
so that $K(100)=50$ and $\eta(100)=0.006$. Thus, as $d$ increases, AFHMC uses longer trajectories but smaller leapfrog steps. We run AFHMC until $\mathcal W_2^2<0.1$ for 20 consecutive checks and record the first communication round at which the threshold is crossed. The 2-Wassertein distance calculation is based on $M(d)=100d(d-1)$ independent parallel chains (note that we need to estimate a $d$-dimensional covariance matrix).

As a matched-budget baseline, we run FA-HMC with the same $\eta(d)$ and the same number of communication rounds as AFHMC. FA-HMC uses $T=10$ local HMC update, and the leapfrog step count within each HMC update is chosen as
\[
  K_{\mathrm{F}}(d)
  =
  \max\left\{1,\left\lfloor K(d)/10 \right\rceil\right\},
\]
so that the total number of leapfrog updates per communication round is matched to AFHMC up to rounding. \revise{To check that the gap below is not caused by giving FA-HMC the AFHMC stepsize, we additionally run FA-HMC for $2R(d)$ rounds, i.e., the same number of server exchanges as AFHMC at round $R(d)$, under two FA-HMC-specific schedules: the dimension-scaling setting of \cite{liang2025bayesian}, $K_{\mathrm{F}}=5$, $T=10$ and $\eta(d)=0.02d^{-1/4}$; and the stepsize suggested by Theorem~4.1 of \cite{liang2025bayesian}, $\eta(d)=0.006(d/100)^{-7/12}$, with $T=10$ and $K_{\mathrm{F}}(d)$ as above (see Appendix~\ref{app:gaussian-extra}).}
\begin{figure}[t]
  \centering
  \captionsetup[figure]{skip=0pt}
  \includegraphics[width=0.32\linewidth]{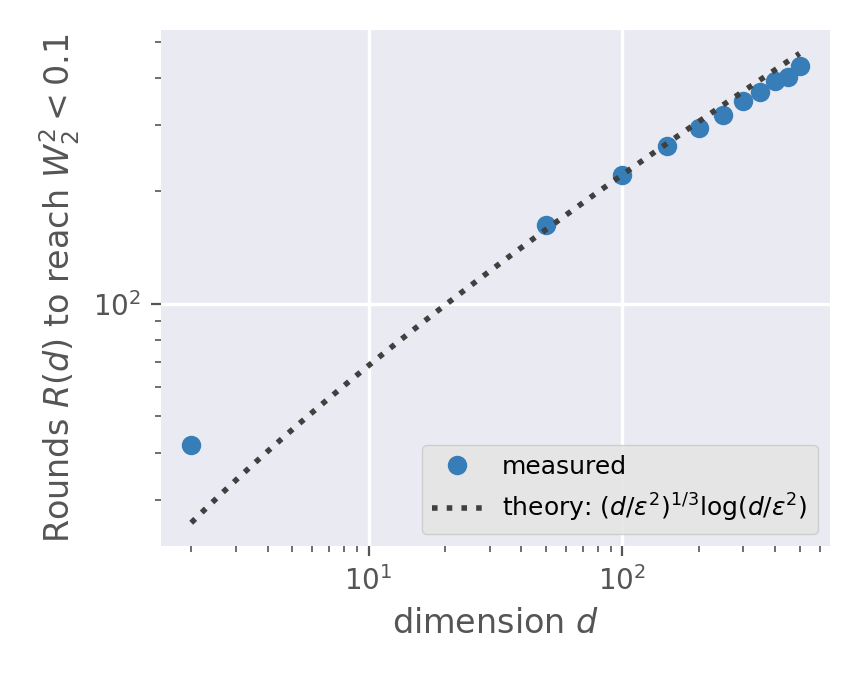}
  \hfill
  \includegraphics[width=0.32\linewidth]{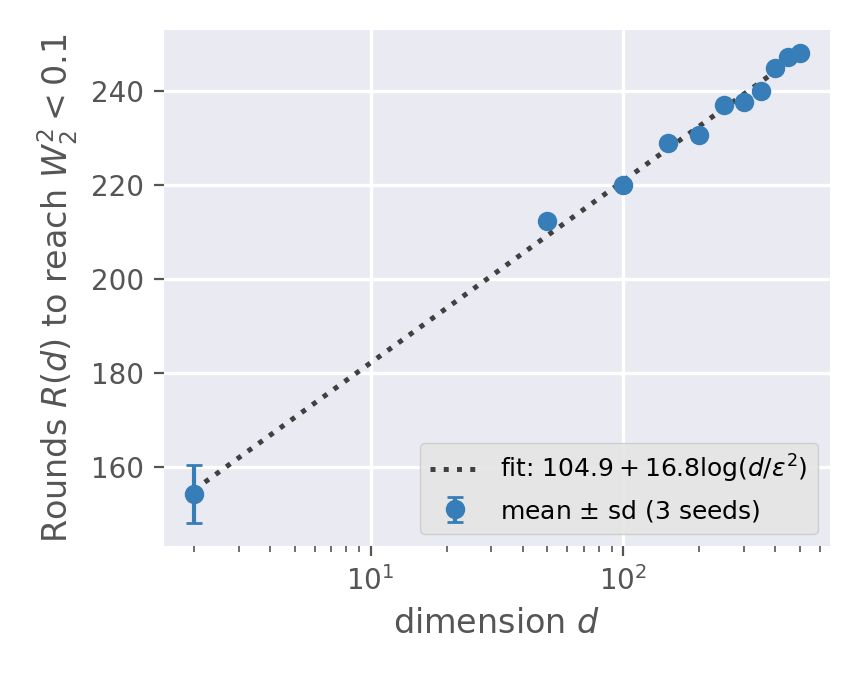}
  \hfill
  \includegraphics[width=0.32\linewidth]{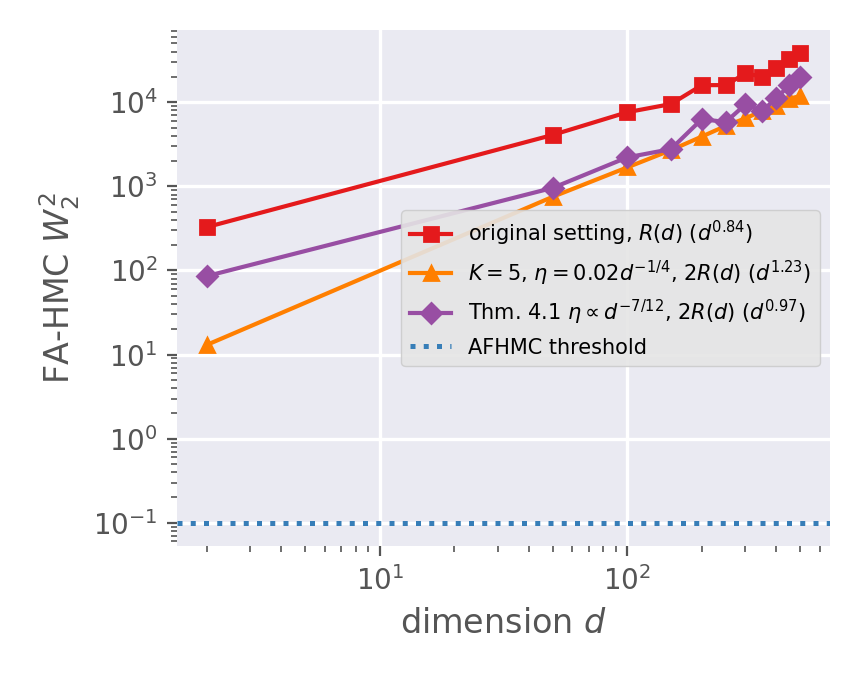}

  \caption{Dimension scaling on Gaussian targets. Left: communication rounds required by AFHMC to reach ${\cal W}_2^2<0.1$ \revise{on the heterogeneous target} as dimension increases. \revise{Middle: the same quantity on the low-heterogeneity target (mean $\pm$ sd over three seeds), with the fitted logarithmic law. Right: FA-HMC ${\cal W}_2^2$ on the heterogeneous target under the original setting ($R(d)$ rounds) and under two FA-HMC-specific stepsize schedules at the equal server-exchange budget $2R(d)$}; the horizontal line is the AFHMC stopping threshold $\mathcal W_2^2=0.1$.}
  \label{fig:gaussian-scaling}
\end{figure}

Figure~\ref{fig:gaussian-scaling} summarizes the dimension-scaling experiment. The left panel shows that the number of communication rounds required by AFHMC increases from $42$ at $d=2$ to $430$ at $d=500$. As shown by the figure, the empirical behavior closely resembles the theoretical trend illustrated by the dotted black curve. When fitting the plot by regression $C (d/\epsilon^2)^\alpha \log(d/\epsilon^2)$ with $\epsilon^2=0.1$ (excluding the outlier case $d=2$), we have $\hat\alpha=0.282$. This log-corrected scaling is close to the predicted $O((d/\epsilon^2)^{1/3}\log(d/\epsilon^2))$ communication complexity, especially on the larger-dimensional regime. The right panel reports the ${\cal W}_2^2$ error of FA-HMC when FA-HMC spends the same dimension-dependent budget used by AFHMC\revise{, together with the two FA-HMC-specific schedules at the equal server-exchange budget}. The horizontal ``adaptive threshold'' line marks ${\cal W}_2^2=0.1$, the stopping accuracy used for AFHMC in \revise{the left panel}. Under the matched budget, FA-HMC remains orders of magnitude above this threshold: its final ${\cal W}_2^2$ ranges from $3.28\times 10^2$ at $d=2$ to $3.81\times 10^4$ at $d=500$, with empirical growth speed $O(d^{0.843})$. \revise{The FA-HMC-specific schedules do not close the gap even with twice as many rounds: at $d=500$, they reach ${\cal W}_2^2=1.19\times 10^4$ and $1.98\times 10^4$ after $860$ rounds, whereas AFHMC crosses $0.1$ at round $430$, and their errors grow as $O(d^{1.23})$ and $O(d^{0.97})$, i.e., roughly linearly in $d$, as expected from a heterogeneity bias.} This significant performance gap between AFHMC and FA-HMC indicates that the FA-HMC dynamics retain a persistent heterogeneity bias, whereas AFHMC reaches a higher level of accuracy and efficiency.

\revise{\paragraph{Low-heterogeneity regime.}%
To examine the $O(\log(d/\epsilon^2))$ communication complexity predicted by Theorem~\ref{thm:convergrate:sg} when the heterogeneity is small, we keep $N=10$ clients but estimate every local Gaussian from $10^5$ samples drawn from the same distribution $\mathcal{N}(16.2\mathbf{1}_d,1.6I_d)$, so that the heterogeneity only comes from sampling error; the pooled samples define the target. We use $K(d)=\lfloor 50(d/100)^{1/2}\rceil$ and $\eta(d)=0.006(d/100)^{-1/2}$, and repeat each setting with three random seeds. AFHMC reaches $\mathcal{W}_2^2<0.1$ for every dimension and seed (Figure~\ref{fig:gaussian-scaling}, middle), and the first-crossing round is well described by $R(d)=104.9+16.8\log(d/\epsilon^2)$ with $R^2=0.996$. As $d$ grows from $50$ to $500$, $R(d)$ only increases from $212$ to $248$ (a factor of $1.17$), compared with an increase from $162$ to $430$ (a factor of $2.65$) on the heterogeneous target. This provides direct numerical evidence for the logarithmic communication complexity in the low-heterogeneity regime (details in Appendix~\ref{app:gaussian-extra}).}

\paragraph{Logistic regression for FMNIST.}

We next evaluate the proposed method on a Bayesian multinomial logistic regression task using the Fashion-MNIST (FMNIST) dataset
\citep{xiao2017fashion}, which consists of $60{,}000$ training images and $10{,}000$ test images from $10$ clothing categories. The dataset is released by Zalando Research under the MIT license. Each image is a $28\times 28$ grayscale image, flattened into a $784$-dimensional feature vector. To model client heterogeneity, we partition the training data across $N=10$ clients using a label-skewed Dirichlet split with concentration parameter $\alpha=0.1$, see Appendix~\ref{app:fmnist-extra} for details.

We compare FA-LD, FA-HMC, and AFHMC under the same heterogeneous data split. For FA-LD we use the $K=1$ case, while for FA-HMC and AFHMC we evaluate $K\in\{10,50,100\}$. Each method is run for \revise{$3600$ server exchanges, i.e., $3600$ rounds for FA-LD and FA-HMC and $1800$ rounds for AFHMC. Between two synchronizations, each client performs $50$ leapfrog steps in FA-LD and FA-HMC and $K$ leapfrog steps in AFHMC.} For each reported metric, we select the stepsize separately for every method and $K$ value.

\begin{wrapfigure}{r}{0.52\linewidth}
  \centering
  \captionsetup[figure]{skip=0pt}
  \includegraphics[width=\linewidth]{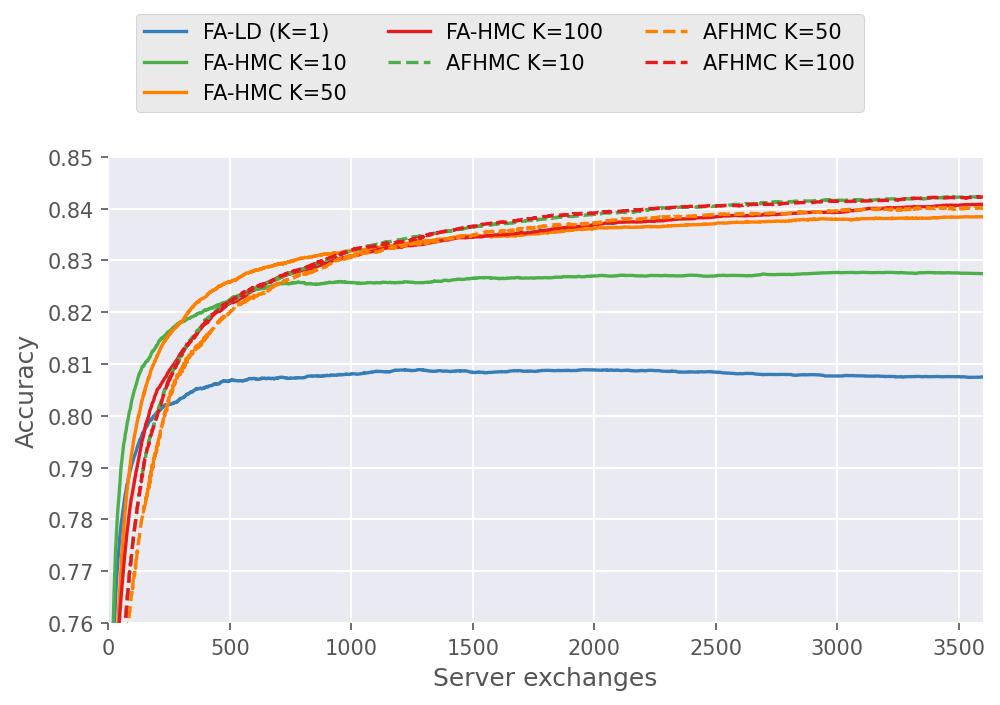}
  \caption{Test accuracy for Bayesian logistic regression on heterogeneous FMNIST \revise{against the number of server exchanges (one per round for FA-LD and FA-HMC, two per round for AFHMC).}}
  \label{fig:fmnist-accuracy}
\end{wrapfigure}

Figure~\ref{fig:fmnist-accuracy} reports the test accuracy \revise{against the number of server exchanges. AFHMC with $K\in\{10,100\}$ overtakes all FA-HMC variants after about $1000$ exchanges, stays ahead thereafter, and attains the highest final accuracy ($0.842$, versus $0.841$ for the best FA-HMC at the same number of exchanges and $0.836$ at the same number of rounds). With $K=10$, AFHMC achieves this using about ten times fewer gradient evaluations per client than FA-HMC.} This \revise{result} shows that the drift correction in AFHMC continues to be useful beyond the Gaussian experiments and transfers to a non-Gaussian Bayesian logistic regression posterior with real image data. Additional predictive-quality metrics, including Brier score and negative log-likelihood, are reported in Appendix~\ref{app:fmnist-extra}.

\section{Conclusions and Discussions}
In this work, we propose a novel Bayesian Federated sampling algorithm based on the Hamiltonian dynamics.  We incorporate a global drift control variate into the leapfrog integrator to alleviate the local drifting caused by the data heterogeneity.
This surrogate leapfrog update rule leads to a significant advancement of the theoretical behavior, especially for the communication cost with respect to the system's high dimensionality.
Our analysis shows that to achieve an $\epsilon$ sampling accuracy, AFHMC only requires $O((d/\epsilon^2)^{1/3})$ communication rounds. Compared with the $O((d/\epsilon^2)^{1/2})$ rate achieved by Langevin-based federated sampler \citep{plassier2023federated, Wei2024On}, our result reaffirms the superiority of Hamiltonian dynamics over Langevin dynamics for the Federated sampling task, promoting the communities to utilize HMC-based Federated learning methods for complicated high-dimensional tasks. The advantage claimed by our theory is justified by empirical evidence.
In addition, AFHMC can achieve $O(\log(d/\epsilon^2))$ rate for communication complexity under low heterogeneity, matching the rate of communication cost for the Federated optimization realm\revise{; our low-heterogeneity Gaussian experiment in Section~\ref{sec:numerical} confirms this logarithmic scaling numerically}.

\noindent\textbf{Limitations.}
The scope of our theoretical investigation is limited to the log-convex target distribution and doesn't consider the stochastic gradient. The theoretical analysis for Hamiltonian dynamics or HMC is highly non-trivial. In the literature, not limited to FL, no result achieves an advanced dimension-dependent rate for HMC under nonconvexity or stochastic gradients. Our contribution is mainly theoretical, hence our numerical experiments are on a small scale, similarly to prior theoretical works, serving for the purpose of validating our theorems.

\begin{ack}
Guang Lin gratefully acknowledges the support from DARPA MATHBAC program under grant number: HR0011-26-3-E152, the National Science Foundation (DMS-2533878, DMS-2053746, DMS-2134209, ECCS-2328241, CBET-2347401 and OAC-2311848), and U.S.~Department of Energy (DOE) Office of Science Advanced Scientific Computing Research program under the "Uncertainty Quantification for Multifidelity Operator Learning (MOLUcQ)" project (Project No. 81739), DE-SC0023161, the SciDAC LEADS Institute, and DOE–Fusion Energy Science, under grant number: DE-SC0024583.

\end{ack}

\bibliography{example_paper}

\begin{thebibliography}{46}
\providecommand{\natexlab}[1]{#1}
\providecommand{\url}[1]{\texttt{#1}}
\expandafter\ifx\csname urlstyle\endcsname\relax
  \providecommand{\doi}[1]{doi: #1}\else
  \providecommand{\doi}{doi: \begingroup \urlstyle{rm}\Url}\fi

\bibitem[Akhmatskaya et~al.(2026)Akhmatskaya, Nagar, Carrillo, {Gavira
  Balmacz}, Inouzhe, {Parga Pazos}, and {Rodríguez
  Álvarez}]{Akhmatskaya2026Adaptive}
Akhmatskaya, E., Nagar, L., Carrillo, J.~A., {Gavira Balmacz}, L., Inouzhe, H.,
  {Parga Pazos}, M., and {Rodríguez Álvarez}, M.~X.
\newblock Adaptive tuning of hamiltonian monte carlo methods.
\newblock \emph{Applied Mathematical Modelling}, 157:\penalty0 116892, 2026.
\newblock ISSN 0307-904X.
\newblock \doi{https://doi.org/10.1016/j.apm.2026.116892}.
\newblock URL
  \url{https://www.sciencedirect.com/science/article/pii/S0307904X26001538}.

\bibitem[Al-Shedivat et~al.(2021)Al-Shedivat, Gillenwater, Xing, and
  Rostamizadeh]{al-shedivat2021federated}
Al-Shedivat, M., Gillenwater, J., Xing, E., and Rostamizadeh, A.
\newblock Federated learning via posterior averaging: A new perspective and
  practical algorithms.
\newblock In \emph{International Conference on Learning Representations}, 2021.
\newblock URL \url{https://openreview.net/forum?id=GFsU8a0sGB}.

\bibitem[Bonawitz et~al.(2019)Bonawitz, Eichner, Grieskamp, Huba, Ingerman,
  Ivanov, Kiddon, Kone{\v{c}}n{\`y}, Mazzocchi, and
  McMahan]{bonawitz2019towards}
Bonawitz, K., Eichner, H., Grieskamp, W., Huba, D., Ingerman, A., Ivanov, V.,
  Kiddon, C., Kone{\v{c}}n{\`y}, J., Mazzocchi, S., and McMahan, B.
\newblock {Towards Federated Learning at Scale: System design}.
\newblock \emph{Proceedings of Machine Learning and Systems}, 1:\penalty0
  374--388, 2019.

\bibitem[Chen et~al.(2022)Chen, Xiao, and Pang]{chen2022satellite}
Chen, H., Xiao, M., and Pang, Z.
\newblock Satellite-based computing networks with federated learning.
\newblock \emph{IEEE Wireless Communications}, 29\penalty0 (1):\penalty0
  78--84, 2022.

\bibitem[Chen \& Chao(2021)Chen and Chao]{chen2021fedbe}
Chen, H.-Y. and Chao, W.-L.
\newblock Fed{\{}be{\}}: Making bayesian model ensemble applicable to federated
  learning.
\newblock In \emph{International Conference on Learning Representations}, 2021.
\newblock URL \url{https://openreview.net/forum?id=dgtpE6gKjHn}.

\bibitem[Chen et~al.(2014)Chen, Fox, and Guestrin]{chen2014stochastic}
Chen, T., Fox, E., and Guestrin, C.
\newblock {Stochastic Gradient Hamiltonian Monte Carlo}.
\newblock In \emph{International Conference on Machine Learning (ICML)}, pp.\
  1683--1691, 2014.

\bibitem[Dalalyan(2017)]{dk17}
Dalalyan, A.~S.
\newblock Theoretical {G}uarantees for {A}pproximate {S}ampling from {S}mooth
  and {L}og-concave {D}ensities.
\newblock \emph{Journal of the Royal Statistical Society: Series B},
  79\penalty0 (3):\penalty0 651--676, 2017.

\bibitem[Dalalyan \& Karagulyan(2019)Dalalyan and Karagulyan]{dk19}
Dalalyan, A.~S. and Karagulyan, A.
\newblock User-friendly {G}uarantees for the {L}angevin {M}onte {C}arlo with
  {I}naccurate {G}radient.
\newblock \emph{Stochastic Processes and their Applications}, 129\penalty0
  (12):\penalty0 5278--5311, 2019.

\bibitem[Deng et~al.(2024)Deng, Zhang, Ma, Song, and Lin]{Wei2024On}
Deng, W., Zhang, Q., Ma, Y.-A., Song, Z., and Lin, G.
\newblock On convergence of federated averaging langevin dynamics.
\newblock In \emph{Proceedings of the Fortieth Conference on Uncertainty in
  Artificial Intelligence}, UAI '24, 2024.

\bibitem[Duane et~al.(1987)Duane, Kennedy, Pendleton, and Roweth]{Duane87}
Duane, S., Kennedy, A., Pendleton, B., and Roweth.
\newblock Hybrid {M}onte {C}arlo.
\newblock \emph{Physics Letters B}, 195\penalty0 (2):\penalty0 216--222, 1987.

\bibitem[Durmus \& Moulines(2019)Durmus and Moulines]{durmus2019high}
Durmus, A. and Moulines, E.
\newblock {High-dimensional Bayesian Inference via The Unadjusted Langevin
  Algorithm}.
\newblock \emph{Bernoulli}, 25\penalty0 (4A):\penalty0 2854--2882, 2019.

\bibitem[Durmus et~al.(2020)Durmus, \'{E}ric Moulines, and
  Saksman]{Irreducibility_HMC}
Durmus, A., \'{E}ric Moulines, and Saksman, E.
\newblock {Irreducibility and Geometric Ergodicity of Hamiltonian Monte Carlo}.
\newblock \emph{Annals of Statistics}, 48\penalty0 (6):\penalty0 3545–3564,
  2020.

\bibitem[Erdogdu \& Hosseinzadeh(2021)Erdogdu and
  Hosseinzadeh]{erdogdu2021convergence}
Erdogdu, M.~A. and Hosseinzadeh, R.
\newblock {On the Convergence of Langevin Monte Carlo: the Interplay between
  Tail Growth and Smoothness}.
\newblock In \emph{Proc. of Conference on Learning Theory (COLT)}, pp.\
  1776--1822, 2021.

\bibitem[Fern{\'a}ndez-Pend{\'a}s et~al.(2016)Fern{\'a}ndez-Pend{\'a}s,
  Akhmatskaya, and Sanz-Serna]{fernandez2016adaptive}
Fern{\'a}ndez-Pend{\'a}s, M., Akhmatskaya, E., and Sanz-Serna, J.~M.
\newblock Adaptive multi-stage integrators for optimal energy conservation in
  molecular simulations.
\newblock \emph{Journal of Computational Physics}, 327:\penalty0 434--449,
  2016.

\bibitem[Girolami \& Calderhead(2011)Girolami and
  Calderhead]{girolami2011riemann}
Girolami, M. and Calderhead, B.
\newblock {Riemann Manifold Langevin and Hamiltonian Monte Carlo Methods}.
\newblock \emph{Journal of the Royal Statistical Society: Series B (Statistical
  Methodology)}, 73\penalty0 (2):\penalty0 123--214, 2011.

\bibitem[Guo et~al.(2023)Guo, Greengard, Wang, Gelman, Kim, and
  Xing]{guo2023federated}
Guo, H., Greengard, P., Wang, H., Gelman, A., Kim, Y., and Xing, E.
\newblock Federated learning as variational inference: A scalable expectation
  propagation approach.
\newblock In \emph{The Eleventh International Conference on Learning
  Representations}, 2023.
\newblock URL \url{https://openreview.net/forum?id=dZrQR7OR11}.

\bibitem[Hirt et~al.(2021)Hirt, Titsias, and Dellaportas]{hirt2021entropy}
Hirt, M., Titsias, M., and Dellaportas, P.
\newblock Entropy-based adaptive hamiltonian monte carlo.
\newblock \emph{Advances in neural information processing systems},
  34:\penalty0 28482--28495, 2021.

\bibitem[Hoffman et~al.(2021)Hoffman, Radul, and Sountsov]{hoffman2021adaptive}
Hoffman, M., Radul, A., and Sountsov, P.
\newblock An adaptive-mcmc scheme for setting trajectory lengths in hamiltonian
  monte carlo.
\newblock In \emph{International Conference on Artificial Intelligence and
  Statistics}, pp.\  3907--3915. PMLR, 2021.

\bibitem[Hoffman \& Gelman(2014)Hoffman and Gelman]{no_U_turn_sampler}
Hoffman, M.~D. and Gelman, A.
\newblock {The No-U-Turn Sampler: Adaptively Setting Path Lengths in
  Hamiltonian Monte Carlo}.
\newblock \emph{Journal of Machine Learning Research}, pp.\  1593--1623, 2014.

\bibitem[Karimireddy et~al.(2020)Karimireddy, Kale, Mohri, Reddi, Stich, and
  Suresh]{kkm+20}
Karimireddy, S.~P., Kale, S., Mohri, M., Reddi, S., Stich, S., and Suresh,
  A.~T.
\newblock Scaffold: Stochastic {C}ontrolled {A}veraging for {F}ederated
  {L}earning.
\newblock In \emph{International Conference on Machine Learning (ICML)}, pp.\
  5132--5143, 2020.

\bibitem[Kim \& Hospedales(2025)Kim and Hospedales]{kim2025fedhb}
Kim, M. and Hospedales, T.
\newblock Fedhb: Hierarchical bayesian federated learning.
\newblock \emph{Journal of Machine Learning Research}, 26\penalty0
  (272):\penalty0 1--50, 2025.

\bibitem[Kone{\v{c}}n{\`y} et~al.(2015)Kone{\v{c}}n{\`y}, McMahan, and
  Ramage]{konevcny2015federated}
Kone{\v{c}}n{\`y}, J., McMahan, B., and Ramage, D.
\newblock {Federated optimization: Distributed optimization Beyond the
  Datacenter}.
\newblock \emph{arXiv preprint arXiv:1511.03575}, 2015.

\bibitem[Kone{\v{c}}n{\`y} et~al.(2016)Kone{\v{c}}n{\`y}, McMahan, Yu,
  Richtarik, Suresh, and Bacon]{konevcny2016federated}
Kone{\v{c}}n{\`y}, J., McMahan, H.~B., Yu, F.~X., Richtarik, P., Suresh, A.~T.,
  and Bacon, D.
\newblock {Federated Learning: Strategies for Improving Communication
  Efficiency}.
\newblock In \emph{NIPS Workshop on Private Multi-Party Machine Learning},
  2016.
\newblock URL \url{https://arxiv.org/abs/1610.05492}.

\bibitem[Li et~al.(2020{\natexlab{a}})Li, Sahu, Talwalkar, and
  Smith]{li2020federated}
Li, T., Sahu, A.~K., Talwalkar, A., and Smith, V.
\newblock {Federated Learning: Challenges, Methods, and Future Directions}.
\newblock \emph{IEEE Signal Processing Magazine}, 37\penalty0 (3):\penalty0
  50--60, 2020{\natexlab{a}}.

\bibitem[Li et~al.(2020{\natexlab{b}})Li, Sahu, Zaheer, Sanjabi, Talwalkar, and
  Smithy]{LS20}
Li, T., Sahu, A.~K., Zaheer, M., Sanjabi, M., Talwalkar, A., and Smithy, V.
\newblock Federated {O}ptimization in {H}eterogeneous {N}etworks.
\newblock In \emph{Proceedings of the 3rd MLSys Conference},
  2020{\natexlab{b}}.

\bibitem[Liang et~al.(2025)Liang, Zhang, Deng, Song, and
  Lin]{liang2025bayesian}
Liang, J., Zhang, Q., Deng, W., Song, Q., and Lin, G.
\newblock Bayesian federated learning with hamiltonian monte carlo: Algorithm
  and theory.
\newblock \emph{Journal of Computational and Graphical Statistics}, 34\penalty0
  (2):\penalty0 509--518, 2025.

\bibitem[Liu et~al.(2024)Liu, Jiang, Zheng, Chen, Qi, Huang, and
  Shao]{LiuBFL2024}
Liu, L., Jiang, X., Zheng, F., Chen, H., Qi, G.-J., Huang, H., and Shao, L.
\newblock A bayesian federated learning framework with online laplace
  approximation.
\newblock \emph{IEEE Transactions on Pattern Analysis and Machine
  Intelligence}, 46\penalty0 (1):\penalty0 1--16, 2024.
\newblock \doi{10.1109/TPAMI.2023.3322743}.

\bibitem[Mangoubi \& Smith(2017)Mangoubi and Smith]{mangoubi2017rapid}
Mangoubi, O. and Smith, A.
\newblock Rapid mixing of hamiltonian monte carlo on strongly log-concave
  distributions.
\newblock \emph{arXiv preprint arXiv:1708.07114}, 2017.

\bibitem[Mangoubi \& Smith(2021)Mangoubi and Smith]{mangoubi2021mixing}
Mangoubi, O. and Smith, A.
\newblock Mixing of hamiltonian monte carlo on strongly log-concave
  distributions: Continuous dynamics.
\newblock \emph{The Annals of Applied Probability}, 31\penalty0 (5):\penalty0
  2019--2045, 2021.

\bibitem[Mangoubi \& Vishnoi(2018)Mangoubi and Vishnoi]{Mangoubi18_leapfrog}
Mangoubi, O. and Vishnoi, N.
\newblock Dimensionally tight bounds for second-order hamiltonian monte carlo.
\newblock In Bengio, S., Wallach, H., Larochelle, H., Grauman, K.,
  Cesa-Bianchi, N., and Garnett, R. (eds.), \emph{Advances in Neural
  Information Processing Systems}, volume~31. Curran Associates, Inc., 2018.
\newblock URL
  \url{https://proceedings.neurips.cc/paper_files/paper/2018/file/e07bceab69529b0f0b43625953fbf2a0-Paper.pdf}.

\bibitem[McMahan et~al.(2017)McMahan, Moore, Ramage, Hampson, and
  y~Arcas]{mmr+17}
McMahan, B., Moore, E., Ramage, D., Hampson, S., and y~Arcas, B.~A.
\newblock {Communication-efficient Learning of Deep Networks from Decentralized
  Data}.
\newblock In \emph{Proc. of the International Conference on Artificial
  Intelligence and Statistics (AISTATS)}, pp.\  1273--1282, 2017.

\bibitem[Mishchenko et~al.(2022)Mishchenko, Malinovsky, Stich, and
  Richtarik]{pmlr-v162-mishchenko22b}
Mishchenko, K., Malinovsky, G., Stich, S., and Richtarik, P.
\newblock {P}rox{S}kip: Yes! {L}ocal gradient steps provably lead to
  communication acceleration! {F}inally!
\newblock In Chaudhuri, K., Jegelka, S., Song, L., Szepesvari, C., Niu, G., and
  Sabato, S. (eds.), \emph{Proceedings of the 39th International Conference on
  Machine Learning}, volume 162 of \emph{Proceedings of Machine Learning
  Research}, pp.\  15750--15769. PMLR, 17--23 Jul 2022.
\newblock URL \url{https://proceedings.mlr.press/v162/mishchenko22b.html}.

\bibitem[Mitra et~al.(2021)Mitra, Jaafar, Pappas, and Hassani]{mitra2021linear}
Mitra, A., Jaafar, R., Pappas, G.~J., and Hassani, H.
\newblock Linear convergence in federated learning: Tackling client
  heterogeneity and sparse gradients.
\newblock \emph{Advances in Neural Information Processing Systems},
  34:\penalty0 14606--14619, 2021.

\bibitem[Nagar et~al.(2024)Nagar, Fern{\'a}ndez-Pend{\'a}s, Sanz-Serna, and
  Akhmatskaya]{nagar2024adaptive}
Nagar, L., Fern{\'a}ndez-Pend{\'a}s, M., Sanz-Serna, J.~M., and Akhmatskaya, E.
\newblock Adaptive multi-stage integration schemes for hamiltonian monte carlo.
\newblock \emph{Journal of Computational Physics}, 502:\penalty0 112800, 2024.

\bibitem[Neal(2012)]{Neal12}
Neal, R.~M.
\newblock {MCMC} using {Hamiltonian} dynamics.
\newblock In \emph{Handbook of Markov Chain {M}onte {C}arlo}, volume~54, pp.\
  113--162. Chapman and Hall/CRC, 2012.

\bibitem[Plassier et~al.(2023)Plassier, Moulines, and
  Durmus]{plassier2023federated}
Plassier, V., Moulines, E., and Durmus, A.
\newblock Federated averaging langevin dynamics: Toward a unified theory and
  new algorithms.
\newblock In \emph{International Conference on Artificial Intelligence and
  Statistics}, pp.\  5299--5356. PMLR, 2023.

\bibitem[Sato \& Nakagawa(2014)Sato and Nakagawa]{sato2014approximation}
Sato, I. and Nakagawa, H.
\newblock Approximation analysis of stochastic gradient langevin dynamics by
  using fokker-planck equation and ito process.
\newblock In \emph{International Conference on Machine Learning}, pp.\
  982--990. PMLR, 2014.

\bibitem[Vedadi et~al.(2024)Vedadi, Dillon, Mansfield, Singhal, Afkanpour, and
  Morningstar]{vedadi2024federated}
Vedadi, E., Dillon, J.~V., Mansfield, P.~A., Singhal, K., Afkanpour, A., and
  Morningstar, W.~R.
\newblock Federated variational inference: Towards improved personalization and
  generalization.
\newblock \emph{Transactions on Machine Learning Research}, 2024.
\newblock ISSN 2835-8856.
\newblock URL \url{https://openreview.net/forum?id=6OmRkUHgs5}.

\bibitem[Wang et~al.(2020)Wang, Yurochkin, Sun, Papailiopoulos, and
  Khazaeni]{wang2020federated}
Wang, H., Yurochkin, M., Sun, Y., Papailiopoulos, D., and Khazaeni, Y.
\newblock {Federated Learning with Matched Averaging}.
\newblock In \emph{International Conference on Learning Representations
  (ICLR)}, 2020.

\bibitem[Wang et~al.(2024)Wang, Chen, Song, and Zhang]{wang2024enhancing}
Wang, Z., Chen, Y., Song, Q., and Zhang, R.
\newblock Enhancing low-precision sampling via stochastic gradient hamiltonian
  monte carlo.
\newblock \emph{Transactions on Machine Learning Research}, 2024.
\newblock ISSN 2835-8856.
\newblock URL \url{https://openreview.net/forum?id=uSLNzzuiDJ}.

\bibitem[Welling \& Teh(2011)Welling and Teh]{Welling11}
Welling, M. and Teh, Y.~W.
\newblock {B}ayesian {L}earning via {S}tochastic {G}radient {L}angevin
  {D}ynamics.
\newblock In \emph{International Conference on Machine Learning (ICML)}, 2011.

\bibitem[Xiao et~al.(2017)Xiao, Rasul, and Vollgraf]{xiao2017fashion}
Xiao, H., Rasul, K., and Vollgraf, R.
\newblock Fashion-mnist: a novel image dataset for benchmarking machine
  learning algorithms.
\newblock \emph{arXiv preprint arXiv:1708.07747}, 2017.

\bibitem[Zhang et~al.(2022)Zhang, Li, Li, Guo, and Shao]{zhang2022personalized}
Zhang, X., Li, Y., Li, W., Guo, K., and Shao, Y.
\newblock Personalized federated learning via variational bayesian inference.
\newblock In \emph{International Conference on Machine Learning}, pp.\
  26293--26310. PMLR, 2022.

\bibitem[Zongchen~Chen(2019)]{Chen_Vempala}
Zongchen~Chen, S. S.~V.
\newblock Optimal {C}onvergence {R}ate of {H}amiltonian {M}onte {C}arlo for
  {S}trongly {L}ogconcave {D}istributions.
\newblock In \emph{Approximation, Randomization, and Combinatorial
  Optimization}, 2019.

\bibitem[Zou \& Gu(2021)Zou and Gu]{zou2021convergence}
Zou, D. and Gu, Q.
\newblock {On the Convergence of Hamiltonian Monte Carlo with Stochastic
  Gradients}.
\newblock In \emph{International Conference on Machine Learning (ICML)}, pp.\
  13012--13022, 2021.

\bibitem[Zou et~al.(2019)Zou, Xu, and Gu]{zou2019stochastic}
Zou, D., Xu, P., and Gu, Q.
\newblock {Stochastic Gradient Hamiltonian Monte Carlo Methods with Recursive
  Variance Reduction}.
\newblock \emph{Advances in Neural Information Processing Systems (NeurIPS)},
  32, 2019.

\end{thebibliography}
\bibliographystyle{icml2024}

\clearpage
\appendix
\section*{Appendix}
\section{Algorithm, assumption and notations}

\subsection{notations and definitions}\label{sec:a}
The ideal HMC dynamic for the global loss $f$ is defined as:
\begin{equation}\label{eq:hmcdynamic}
       d\theta(t)=p(t)dt,\qquad dp(t)=-\nabla_\theta f(\theta(t))dt.
\end{equation}

According to Algorithm~\ref{alg:AFAHMC}, we define $\theta^{(c)}_{t,j}$ and $p^{(c)}_{t,j}$ to be the local position variable and local momentum variable at the $j$-th surrogate leapfrog step of the $c$-th local node within the $t$-th global iteration. In particular, $p^{(c)}_{t,0}=p_t$ is a freshly sampled Gaussian variable at iteration $t$ shared by all local nodes; for $j>0$, $p^{(c)}_{t,j}$ is generated following the local surrogate leapfrog. Also, we have $\theta_t:=\sum_{c=1}^Nw_c \theta^{(c)}_{t-1,K}$ aggregated from local nodes and $\theta^{(c)}_{t,0}:=\theta_t$ broadcast from the center node.

Define $\theta_{t,k}:=\sum_{c=1}^Nw_c \theta^{(c)}_{t,j}$ as the corresponding (implicit) global position variable, and $p_{t,k}=\sum_{c=1}^Nw_c p^{(c)}_{t,j}$ is the corresponding (implicit) global momentum variable.

Define ${\cal H}_\eta(\theta):=\theta+\eta p-(\eta^2/2)\nabla f(\theta)$  as the classical discrete leapfrog update from position $\theta$ associated with momentum $p$ and stepsize $\eta$ and $\barHe(\theta):=\theta+\int_0^\eta p(t)dt$ as the classical continuous leapfrog update from position $\theta$ associated with momentum $p(t)$ and length $\eta$, following the HMC dynamic (\ref{eq:hmcdynamic}).
Similarly, define ${\cal H}_\eta(p):=p-\eta/2(\nabla f(\theta)+\nabla f(\theta+\eta p-\eta^2/2\nabla f(\theta)))$ as a  classical discrete leapfrog update from $p$ associated with position $\theta$ and stepsize $\eta$ and $\barHe(p):=p-\int_0^\eta \theta(t)dt$ as a classical continuous leapfrog update from $p$ associated with position $\theta(t)$ and length $\eta$ following the HMC dynamic (\ref{eq:hmcdynamic}).

Based on these, we use the shorthands ${\cal H}^{1}_\eta (\theta):={\cal H}_\eta (\theta)$, ${\cal H}_\eta^{i+1}(\theta):={\cal H}_\eta ({\cal H}^{i}_\eta (\theta))={\cal H}_\eta^i(\theta)+\eta {\cal H}_\eta^i(p)-\eta^2/2\nabla f({\cal H}_\eta^i(\theta))$, and ${\cal H}^{1}_\eta (p):={\cal H}_\eta (p)$, ${\cal H}_\eta^{i+1}(p):={\cal H}_\eta ({\cal H}^{i}_\eta (p))={\cal H}_\eta^i(p)-\eta^2/2(\nabla f({\cal H}_\eta^i(\theta))+\nabla f({\cal H}_\eta^i(\theta)+\eta {\cal H}_\eta^i(p)-\eta^2/2\nabla f({\cal H}_\eta^i(\theta))))$. We also use the shorthands
$\barHe^i(\theta)=\theta+\int_0^{i\eta} p(s)ds$ and $\barHe^i(p):=p-\int_0^{i\eta} \theta(s)ds$.

In addition, we define a coupled continuous HMC process as follows.  $\theta^\pi_0$ follows  the target distribution $\propto \exp(-f)$. $\theta^\pi_{t,0}:=\theta_t^\pi$, $\theta_{t,k+1}^\pi:=\bar{{\cal H}}_{\eta}(\theta^\pi_{t,k})$, and
$\theta_{t+1}^\pi:=\theta_{t,K}^\pi=\bar{{\cal H}}_{K\eta}(\theta^\pi_{t,0})$. For this process, the associated initial momentum is the same as the discrete HMC process, i.e., $p^\pi_{t,0}=p_t^\pi= p_{t}$.

We also denote $\theta^*=\text{argmin}_{\theta}f(\theta)$.

\section{Proofs}\label{sec:B}

\subsection{main proof}
\thmconvergratesg*
\begin{proof}

Let us define
\begin{equation}\label{eq:thmconvergratesg:2.5}
    K^2\eta^2 = C(\frac{\epsilon^2m^2}{HM^2})^{1/3},\qquad \eta^2 =C\frac{\epsilon^2m^2}{M^2d}
\end{equation} for some sufficiently small $C$,
 we have
\begin{equation}\label{eq:thmconvergratesg:3}
   \frac{K^2\eta^4\frac{M^2}{m}d}{mK^2\eta^2}= C \epsilon^2\leq \frac{1}{3}\epsilon^2, \qquad \frac{(K\eta)^8\frac{M^2}{m}H}{mK^2\eta^2}= C^3\epsilon^2\leq \frac{1}{3}\epsilon^2.
\end{equation}
    One can check that if \[
    t \geq \frac{\log(\epsilon^2/ (3\mathbb{E}\|\theta_0-\theta^\pi_0\|^2))}{\log(1-Cm(\frac{\epsilon^2m^2}{HM^2})^{1/3}/16)}\asymp \frac{1}{C}(\frac{HM^2}{\epsilon^2m^5})^{1/3}(\log(3\mathbb{E}\|\theta_0-\theta^\pi_0\|^2/\epsilon^2)),
    \]
    we have
    \begin{equation}\label{eq:thmconvergratesg:4}
    (1-m(K\eta)^2/16)^t \mathbb{E}\|\theta_t-\theta_t^\pi\|^2\leq \frac{1}{3}\epsilon^2.
    \end{equation}
    Combining (\ref{eq:thmconvergratesg:3})-(\ref{eq:thmconvergratesg:4}) and by Theorem~\ref{thm:AFLHMC}, we can see that, for enough small $C$ and set (\ref{eq:thmconvergratesg:2.5}), we can have
\[
\mathbb{E}\|\theta_{t}-\theta_{t}^\pi\|^2\leq \epsilon^2.
\]
This justifies the required communication round
\[t_\epsilon\geq \frac{1}{C}(\frac{HM^2}{\epsilon^2m^5})^{1/3}\log(3\mathbb{E}\|\theta_0-\theta^\pi_0\|^2/\epsilon^2)\]

On the other hand, if $H \ll M^4\epsilon^2/m$ (i.e., the heterogeneity cross local nodes is small), by choosing enough small $C$ such that with $K^2\eta^2 = Cm/M^2$ and $\eta^2=C\epsilon^2 m^2/(M^2d)$, we have
\begin{equation}\label{eq:thmconvergratesg:1}
   \frac{K^2\eta^4\frac{M^2}{m}d}{mK^2\eta^2}\leq C^2 \epsilon^2\leq \frac{1}{3}\epsilon^2, \qquad \frac{(K\eta)^8\frac{M^2}{m}H}{mK^2\eta^2}=C^3\frac{m}{M^4}H\ll \epsilon^2\leq \frac{1}{3}\epsilon^2.
\end{equation}
    One can check that if \[
    t \geq \frac{\log(\epsilon^2/ (3\mathbb{E}\|\theta_0-\theta^\pi_0\|^2))}{\log(1-Cm^2/(16M^2))},
    \]
    we have
    \begin{equation}\label{eq:thmconvergratesg:2}
    (1-\frac{Cm^2}{M^2})^t \mathbb{E}\|\theta_t-\theta_t^\pi\|^2\leq \frac{1}{3}\epsilon^2.
    \end{equation}
Combining (\ref{eq:thmconvergratesg:1})-(\ref{eq:thmconvergratesg:2}) and by Theorem~\ref{thm:AFLHMC}, we obtain that
$\mathbb{E}\|\theta_{t}-\theta_{t}^\pi\|^2\leq \epsilon^2$.
This justified the required communication round to be
\[t_\epsilon\geq C'\log(3\mathbb{E}\|\theta_0-\theta^\pi_0\|^2/\epsilon^2).\]
Finally, we can choose $\theta^\pi_0$ such that ${\cal W}^2(\pi_t,\pi)=\mathbb{E}\|\theta_0-\theta^\pi_0\|^2$, which concludes the proof.
\end{proof}

\thmAFLHMC*
\begin{proof}
We introduce an intermediate variable $ \barHe^K(\theta_t)$ being the continuously updated version of $\theta_{t,K}$ from $\theta_t$. For $c_1,c_2>0$ with $c_1+c_2=1$, by Lemma~\ref{lem:convex}, we can separate the errors into two parts
\begin{equation}\label{pfmain:0:1}
    \|\theta_{t,K}-\theta^\pi_{t,K}\|^2\leq \frac{1}{c_1}\underbrace{\| \theta_{t,K}-\barHe^K(\theta_t)\|^2}_{\text{discrete divergence error}}+\frac{1}{c_2}\underbrace{\|\barHe^K(\theta_t)-\theta^\pi_{t,K}\|^2}_{\text{ideal convergence error}}
\end{equation}
The convergence of the ideal process has been well studied in the literature \cite{Chen_Vempala}. Indeed, recall the definitions that that $\barHe^K(\theta_t)$ and $\theta^\pi_{t,K}$ share the same initial momentum $p_t$ for each $t$,  by Lemma~\ref{lem:oraXiP}, it can be bounded by a contraction factor of the distance between their initial positions
\begin{equation}\label{pfmain:0:2}
    \|\barHe^K(\theta_t)-\theta^\pi_{t,K}\|^2\leq (1-\frac{m(K\eta)^2}{4})\|\theta_t-\theta^\pi_t\|^2,\qquad\text{for } K\eta \leq 1/(2\sqrt{M}).
\end{equation}
We choose $c_1,c_2$ such that $\frac{1-m(K\eta)^2/4}{c_2}=1-m(K\eta)^2/8$, and it follows
\begin{equation}\label{pfmain:0:3}
c_2=\frac{1-m(K\eta)^2/4}{1-m(K\eta)^2/8},\qquad c_1=\frac{m(K\eta)^2/8}{1-m(K\eta)^2/8}\geq \frac{m(K\eta)^2}{8}
\end{equation}

This choice ensures that, after amplification by $1/c_1$, the contraction factor in (\ref{pfmain:0:1}) is still strictly smaller than 1.

In the following, to finish the proof, it suffice to show that the divergence error
\[
\mathbb{E}\| \theta_{t,K}-\barHe^K(\theta_t)\|^2 \leq{  \frac{m(K\eta)^2}{16}\|\theta_t-\theta^\pi_t\|^2} + C_1K^2\eta^4\frac{M^2}{m}d+C_2(K\eta)^8\frac{M^2}{m}H,
\]
or some constants $C_1,C_2>0$.
To study the discrete divergence error, we separate the error
 within the leapfrog-step level. Since by definitions $\theta_{t,K}=\sum_{c=1}^N\theta^{(c)}_{t,K}$ and $\theta_t=\theta_{t,0}$ we can write
 \[
 \theta_{t,K}-\barHe^K(\theta_t)=\theta_{t,K}-\barHe^K(\theta_{t,0})=\sum_{k=1}^K\barHe^{K-k}(\theta_{t,k})-\barHe^{K-k+1}(\theta_{t,k-1}).
 \]
 By convexity of $\|\cdot\|^2$ and Lemma~\ref{lem:convex},
\[
\begin{split}
\|\theta_{t,K}-\barHe^K(\theta_t)\|^2=&\|\sum_{k=1}^K\barHe^{K-k}(\theta_{t,k})-\barHe^{K-k+1}(\theta_{t,k-1})\|^2\\
\leq& K  \sum_{k=1}^K\|\barHe^{K-k}(\theta_{t,k})-\barHe^{K-k+1}(\theta_{t,k-1})\|^2
\end{split}
\]
We emphasis that $\barHe^{K-k}(\theta_{t,k})$ starts with position $\theta_{t,k}$ and momentum $p_{t,k}$ while $\barHe^{K-k+1}(\theta_{t,k-1})$ can be viewed as a $(K-k)\eta$-length process starting with position $\barHe(\theta_{t,k-1})$ and momentum $\barHe(p_{t,k-1})$.  Lemma~\ref{lem:oraXiP} yields
\[
\|\barHe^{K-k}(\theta_{t,k})-\barHe^{K-k+1}(\theta_{t,k-1})\|^2 \leq 2\|\theta_{t,k}-\barHe(\theta_{t,k-1})\|^2 +2(K-k)^2\eta^2\|p_{t,k}-\barHe(p_{t,k-1})\|^2.
\]
It follows that, we can bound the divergence error as follow
\begin{equation}\label{pfmain:1}
    \|\theta_{t,K}-\barHe^K(\theta_t)\|^2\leq
    2K  \sum_{k=1}^K\Bigl(\underbrace{\|\theta_{t,k}-\barHe(\theta_{t,k-1})\|^2}_{\text{error induced by inaccurate positions}}
    +(K-k)^2\eta^2\underbrace{\|p_{t,k}-\barHe(p_{t,k-1})\|^2}_{\text{error induced by inaccurate momentums}} \Bigr).
\end{equation}
The divergence errors can be further separated into systematic leapfrog approximation error and surrogate-induced error. To see this, by definitions
\begin{align*}
    \theta_{t,k}-\barHe(\theta_{t,k-1})=&\underbrace{\He(\theta_{t,k-1})-\barHe(\theta_{t,k-1})}_{\text{systematic leapfrog gap}}-\frac{\eta^2}{2}\sum_{c=1}^Nw_c(\underbrace{\nabla f^{(c)}(\theta^{(c)}_{t,k-1})-\nabla f^{(c)}(\theta_{t,k-1})}_{\text{surrogate induced gap}})\\
    p_{t,k}-\barHe(p_{t,k-1})=&\underbrace{\He(p_{t,k-1})-\barHe(p_{t,k-1})}_{\text{systematic leapfrog gap}}\\&-\frac{\eta}{2}\sum_{c=1}^Nw_c(\underbrace{\nabla f^{(c)}(\theta^{(c)}_{t,k-1})-\nabla f^{(c)}(\theta_{t,k-1})+\nabla f^{(c)}(\theta^{(c)}_{t,k})-\nabla f^{(c)}(\theta_{t,k})}_{\text{surrogate induced gap}}).
\end{align*}
Applying Lemma~\ref{lem:convex}, we can have
\begin{align*}
     \|\theta_{t,k}-\barHe(\theta_{t,k-1})\|^2\leq &2\|\He(\theta_{t,k-1})-\barHe(\theta_{t,k-1})\|^2+2\frac{\eta^4}{4}\sum_{c=1}^Nw_c\|\nabla f^{(c)}(\theta^{(c)}_{t,k-1})-\nabla f^{(c)}(\theta_{t,k-1})\|^2\\
    \|p_{t,k}-\barHe(p_{t,k-1})\|^2\leq &2\|\He(p_{t,k-1})-\barHe(p_{t,k-1})\|^2+2\frac{\eta^2}{4}\sum_{c=1}^Nw_c\sum_{j=k-1}^k2\|\nabla f^{(c)}(\theta^{(c)}_{t,j})-\nabla f^{(c)}(\theta_{t,j})\|^2.
\end{align*}
Inserting these beck to (\ref{pfmain:1}), we have
\begin{align}
\begin{split}\label{pfmain:2}
    &\|\theta_{t,K}-\barHe^K(\theta_t)\|^2\\
    \leq & 2K\sum_{k=1}^K\Bigl( 2\|\He(\theta_{t,k-1})-\barHe(\theta_{t,k-1})\|^2+2(K-k)^2\eta^2\|\He(p_{t,k-1})-\barHe(p_{t,k-1})\|^2\\
    &+\frac{\eta^4}{2}\sum_{c=1}^Nw_c(\|\nabla f^{(c)}(\theta^{(c)}_{t,k-1})-\nabla f^{(c)}(\theta_{t,k-1})\|^2+(K-k)^2\sum_{j=k-1}^k\|\nabla f^{(c)}(\theta^{(c)}_{t,j})-\nabla f^{(c)}(\theta_{t,j})\|^2)\Bigr)\\
    \leq & 5K\sum_{k=0}^{K-1}\Bigl(\underbrace{ \|\He(\theta_{t,k})-\barHe(\theta_{t,k})\|^2+(K-k)^2\eta^2\|\He(p_{t,k})-\barHe(p_{t,k})\|^2}_{\text{divergence error induced by leapfrog approximation}}\\
    &+(K\eta)^2\eta^2     \underbrace{\sum_{c=1}^Nw_c\|\nabla f^{(c)}(\theta^{(c)}_{t,k})-\nabla f^{(c)}(\theta_{t,k})\|^2}_{\text{divergence error induced by surrogate leapfrog update}}\Bigr),
\end{split}
\end{align}
where we have applied $2\eta^4(1/2+2(K-k)^2)\leq 5K^2\eta^4$ for $k< K$ in the last inequality.

\revise{
By $M$-smoothness assumption,
\[
\sum_{c=1}^Nw_c\mathbb{E}\|\nabla f^{(c)}(\theta^{(c)}_{t,k})-\nabla f^{(c)}(\theta_{t,k})\|^2 \leq\sum_{c=1}^Nw_cM^2\mathbb{E}\|\theta^{(c)}_{t,k}-\theta_{t,k}\|^2
\]
and further by Lemma~\ref{lem:diffB} that $\sum_{c=1}^Nw_c\mathbb{E}\|\theta^{(c)}_{t,k}-\theta_{t,k}\|^2\leq k^3\eta^4 \sum_{i=1}^{k-1}(16M^2\mathbb{E}\|\theta^\pi_{t,i}-\theta_{t,i}\|^2+16M^2\mathbb{E}\|\theta^\pi_t-\theta_t\|^2+4(k\eta)^2H) $, we have
\begin{equation}\label{pfmain:3}
    \sum_{c=1}^Nw_c\mathbb{E}\|\nabla f^{(c)}(\theta^{(c)}_{t,k})-\nabla f^{(c)}(\theta_{t,k})\|^2 \leq k^3\eta^4M^2 \sum_{i=1}^{k-1}(16M^2\mathbb{E}\|\theta^\pi_{t,i}-\theta_{t,i}\|^2+16M^2\mathbb{E}\|\theta^\pi_t-\theta_t\|^2+4(k\eta)^2H).
\end{equation}
Here we recall that $H:=\max_{i\leq K}\sum_{c=1}^Nw_c\mathbb{E}\|\frac{1}{i\eta}(\nabla f^{(c)}(\barHe^i(\theta^\pi))-\nabla f^{(c)}(\theta^\pi)+\nabla f(\theta^\pi)-\nabla f(\barHe^i(\theta^\pi)))\|^2$ is a quantity measuring the federated
 heterogeneity.

As for the systematic leapfrog approximation errors, by Lemma~\ref{lem:PosDivB} and Lemma~\ref{lem:MomDivB},  we have
\begin{align}
\begin{split}\label{pfmain:4}
    \|\He(\theta_{t,k})-\barHe(\theta_{t,k})\|^2\leq & \eta^6\Bigl(\frac{M^2}{10}\|p_{t,k}\|^2+\frac{M^4\eta^2}{60}\|\theta_{t,k}-\theta^*\|^2\Bigr) \\
    \|\He(p_{t,k})-\barHe(p_{t,k})\|^2\leq & \eta^4\Bigl( \frac{5M^2}{2}\|p_{t,k}\|^2+\frac{2\eta^2M^4}{3}\|\theta_{t,k}-\theta^*\|^2\Bigr)
\end{split}
\end{align}

Inserting (\ref{pfmain:3})-(\ref{pfmain:4}) back to (\ref{pfmain:2}) and applying Lemma~\ref{lem:AveLocB} on $p_{t,k}$ and $\theta_{t,k}$, we obtain
\begin{align*}
   \mathbb{E} \|\theta_{t,K}-\barHe^K(\theta_t)\|^2\leq & 5K\sum_{k=0}^{K-1}\Bigl((\frac{\eta^6M^2}{10}+\frac{(K\eta)^2\eta^45M^2}{2})\mathbb{E}\|p_{t,k}\|^2+(\frac{\eta^8M^4}{60}+\frac{(K\eta)^22\eta^6M^4}{3})\mathbb{E}\|\theta_{t,k}-\theta^*\|^2\\
    &+  K^5\eta^8M^2 \sum_{i=1}^{k-1}(16M^2\mathbb{E}\|\theta^\pi_{t,i}-\theta_{t,i}\|^2+16M^2\mathbb{E}\|\theta^\pi_t-\theta_t\|^2+4(k\eta)^2H)\Bigr)\\
    \leq & 5K\sum_{k=0}^{K-1}\Bigl((\frac{\eta^6M^2}{10}+ \frac{(K\eta)^2\eta^45M^2}{2})(5\mathbb{E}|p_t\|^2+5M^2(k\eta)^2\mathbb{E}\|\theta_t-\theta^*\|^2)\\
    &+(\frac{\eta^8M^4}{60}+\frac{(K\eta)^22\eta^6M^4}{3})(4(k\eta)^2\mathbb{E}\|p_t\|^2+3\mathbb{E}\|\theta_t-\theta^*\|^2)\\
    &+ K^5\eta^8M^2 \sum_{i=1}^{k-1}(16M^2\mathbb{E}\|\theta^\pi_{t,i}-\theta_{t,i}\|^2+16M^2\mathbb{E}\|\theta^\pi_t-\theta_t\|^2+4(k\eta)^2H)\Bigr).
\end{align*}

}

{

Rearranging the coefficients, it can be seen that
\begin{align*}
     \mathbb{E} \|\theta_{t,K}-\barHe^K(\theta_t)\|^2\leq & \Bigl(\frac{5}{2}K^2\eta^6M^2+\frac{125}{2}K^4\eta^6M^2+\frac{K^4\eta^{10}M^4}{3}+\frac{40K^6\eta^{10}M^4}{3}\Bigr)\mathbb{E}\|p_t\|^2\\
     &+\Bigl(\frac{5}{2}K^4\eta^8M^4+\frac{125}{2}K^6\eta^8M^4+\frac{K^2\eta^8M^4}{4}+80K^4\eta^8M^4\Bigr)\mathbb{E}\|\theta_t-\theta^*\|^2\\
    &+ 80 \sum_{k=0}^{K-1}K^6\eta^8M^4 \sum_{i=1}^{k-1}\mathbb{E}\|\theta^\pi_{t,i}-\theta_{t,i}\|^2\\
        &+80(K\eta)^8M^4\mathbb{E}\|\theta^\pi_t-\theta_t\|^2\\
            &+20(K\eta)^{10}M^2H\\
            \leq& CK^4\eta^6M^2d+CK^6\eta^8M^4\mathbb{E}\|\theta_t-\theta^*\|^2+CK^7\eta^8M^4\sum_{i=1}^{K-2}\mathbb{E}\|\theta^\pi_{t,i}-\theta_{t,i}\|^2\\
            &+C(K\eta)^8M^4\mathbb{E}\|\theta^\pi_t-\theta_t\|^2+C(K\eta)^{10}M^2H
\end{align*}
for enough large $C$ and $K\eta \leq 1/(2\sqrt{M})$.

Combining this with $\mathbb{E}\|\theta_t-\theta^*\|^2\leq 2 \mathbb{E}\|\theta_t-\theta^\pi_t\|^2+2 \mathbb{E}\|\theta^\pi_t-\theta^*\|^2$ and Lemma~\ref{lem:Oracle} that $\mathbb{E}\|\theta^\pi_t-\theta^*\|^2\leq d/m$,  we get
\begin{align}
\begin{split}
    \mathbb{E} \|\theta_{t,K}-\barHe^K(\theta_t)\|^2
            \leq & C'K^4\eta^6M^2d+C'K^7\eta^8M^4\sum_{i=1}^{K-2}\mathbb{E}\|\theta^\pi_{t,i}-\theta_{t,i}\|^2\\
            &+C'(K\eta)^8M^4\mathbb{E}\|\theta^\pi_t-\theta_t\|^2+C'(K\eta)^{10}M^2H
\end{split}
\end{align}
for enough large $C'$.

Finally,  combining it with (\ref{pfmain:0:1})-(\ref{pfmain:0:3}), it can be seen that for an enough large $C''$
\begin{align}
\begin{split}\label{pfmain:4.51new}
      \mathbb{E} \|\theta_{t,K}-\theta^\pi_{t,K}\|^2\leq &(1-\frac{m(K\eta)^2}{8}+\frac{C''(K\eta)^6M^4}{m})  \mathbb{E}\|\theta_t-\theta^\pi_t\|^2+C''K^2\eta^4\frac{M^2}{m}d\\
      &+C''K^5\eta^6\frac{M^4}{m}\sum_{i=1}^{K-2}\mathbb{E}\|\theta^\pi_{t,i}-\theta_{t,i}\|^2
            +C''(K\eta)^8\frac{M^2}{m}H.
            \end{split}
\end{align}
To prove our final claim of this theorem, we use induction to tackle $\sum_{i=1}^{K-2}\mathbb{E}\|\theta^\pi_{t,i}-\theta_{t,i}\|^2$. In specific, we are going to show that there exist constants $C_0,C_1,C_2$ (to be determined later) such that for $K\eta\leq \sqrt{m}/(C_0M)\leq 1/(2\sqrt{M})$
\begin{equation}\label{pfmain:5new}
    \mathbb{E} \|\theta_{t,K}-\theta^\pi_{t,K}\|^2\leq (1-\frac{m(K\eta)^2}{16})  \mathbb{E}\|\theta_t-\theta^\pi_t\|^2+C_1K^2\eta^4\frac{M^2}{m}d+C_2(K\eta)^8\frac{M^2}{m}H.
\end{equation}

Firstly, (\ref{pfmain:5new}) holds for $K=0,1,2$ by {(\ref{pfmain:4.51new}) for a $C_0$ with $16C_0^4>C''$}.

Next, assume that (\ref{pfmain:5new}) holds for all $K\leq \widetilde{K}-1$, we are going to show that (\ref{pfmain:5new}) also holds for $K=\widetilde{K}$.

Applying our induction assumption on (\ref{pfmain:4.51new}), we have
\begin{align*}
        &\mathbb{E} \|\theta_{t,\widetilde{K}}-\barHe^{\widetilde{K}}(\theta_t)\|^2\\
            \leq &  (1-\frac{m(K\eta)^2}{8}+\frac{C''(K\eta)^6M^4}{m})  \mathbb{E}\|\theta_t-\theta^\pi_t\|^2+C''K^2\eta^4\frac{M^2}{m}d+C''(K\eta)^8\frac{M^2}{m}H\\
      &+C''K^5\eta^6\frac{M^4}{m}\sum_{i=1}^{K-2}\Bigl((1-\frac{m(i\eta)^2}{16})  \mathbb{E}\|\theta_t-\theta^\pi_t\|^2+C_1i^2\eta^4\frac{M^2}{m}d+C_2(i\eta)^8\frac{M^2}{m}H\Bigr)\\
      \leq & (1-\frac{m(K\eta)^2}{8}+\frac{C''(K\eta)^6M^4}{m}+C''(K\eta)^6\frac{M^4}{m})\mathbb{E}\|\theta_t-\theta^\pi_t\|^2\\
      &+(C''+C''K^6\eta^6\frac{M^4}{m}C_1)K^2\eta^4\frac{M^2}{m}d+(C''+C''K^6\eta^6C_2)(K\eta)^8\frac{M^2}{m}H
\end{align*}
To finish the induction, we need to satisfy that for $K\eta \leq \sqrt{m}/(C_0M)$,
\[
\frac{m(K\eta)^2}{8}-\frac{2C''(K\eta)^6M^4}{m}\geq \frac{m(K\eta)^2}{16},\, (C''+C''K^6\eta^6\frac{M^4}{m}C_1)\leq C_1,\,  (C''+C''K^6\eta^6C_2)\leq C_2.
\]
which holds trivally for an enough large constant $C_0$.

Noting that $M\geq m$ and that $\theta_{t+1}=\theta_{t,K}$ and $\theta^\pi_{t+1}=\theta^\pi_{t,K}$, this finishes the proof.

}
\end{proof}

\subsection{ideal process}

\begin{restatable}[Lemma D.3 in~\cite{liang2025bayesian}]{lemma}{lemGuD}\label{lem:oraXiP}
Under Assumption \ref{assum:smooth}  consider two HMC systems $\theta(t)$ and $\bar{\theta}(t)$ starting with the same initial positions and different initial momentum $p(0)$ and $\widetilde{p}(0)$ respectively. We have
\[
\|\theta(t)-\bar{\theta}(t)\|_{2} \leq \frac{1}{\sqrt{M}}\sinh(\sqrt{M}t)\|p(0)-\widetilde{p}(0)\|_{2},
\]
Further if we have Assumption  \ref{assum:convex}, consider two HMC systems $\theta(t)$ and $\widetilde{\theta}(t)$ starting with initial positions  $\theta(0)$ and  $\widetilde{\theta}(0)$, and initial momentum $p(0)$ and $\widetilde{p}(0)$ respectively, then
for $0\leq t\leq \frac{1}{2\sqrt{M}}$, we have
\[
\|\theta(t)-\widetilde{\theta}(t)\|_{2} \leq (1-\frac{m}{4}t^2)\|\theta(0)-\widetilde{\theta}(0)\|_{2}+t\|p(0)-\widetilde{p}(0)\|_{2}.
\]
\end{restatable}

\begin{restatable}[Proposition 1 in \cite{durmus2019high}]{lemma}{lemOracle}\label{lem:Oracle} Let $x^*\in \mathbb{R}^d$ be the global minimizer of a loss function $f(\cdot)$ satisfying Assumption \ref{assum:convex} and $x^{\pi}$ be a random variable following distribution $\propto e^{-f(x)}$, then
\[
\mathbb{E}\|x^*-x^{\pi}\|_2^2\leq \frac{d}{m}.
\]
\end{restatable}
\subsection{discrete error control}

\begin{restatable}[Approximation error on position of classical HMC]{lemma}{lemPosDivB}\label{lem:PosDivB} Suppose $p$ is the associated initial momentum of position $q$.
Under Assumption \ref{assum:smooth},
\[
\|\He(q)-\barHe(q)\|^2\leq \eta^6\Bigl(\frac{M^2}{10}\|p\|^2+\frac{M^4\eta^2}{60}\|q-\theta^*\|^2\Bigr).
\]
\end{restatable}

\begin{proof}
By definitions,
\[
\He(q)-\barHe(q)=q+\eta p-\frac{\eta^2}{2}\nabla f(q)-(q+\int_0^{\eta}p(s)ds),\qquad \text{with }\qquad p(s)=p-\int_0^s\nabla f(q(u))du
\]
Inserting $p(s)=p-\int_0^s\nabla f(q(u))du$ and $\frac{\eta^2}{2}\nabla f(q)=\int_0^{\eta}\int_0^s \nabla f(q)duds$, we get
\[
\He(q)-\barHe(q)=\int_0^{\eta}\int_0^s\nabla f(q(u))-\nabla f(q)duds
\]

By Jensen inequality (i.e., Lemma \ref{lem:convex} with $\lambda\equiv1$) and $M$-smoothness assumption of $\nabla f$,
\begin{equation}\label{pflem:PosDivB:1}
    \|\He(q)-\barHe(q)\|^2\leq \frac{\eta^2}{2}\int_0^{\eta}\int_0^s M^2\| q(u) - q \|^2duds.
\end{equation}

In the following, we are going to apply Lemma~\ref{lem:oraXiP} to bound $\| q(u) - q \|^2$. First, by Cauchy–Schwarz inequality and Lemma \ref{lem:convex}
\begin{align*}
    \| q(u) - q \|^2=&\|up-\int_0^{u}\int_0^s \nabla f(q(t))dtds\|^2
    \leq 2\|up\|^2+2\|\int_0^{u}\int_0^s\nabla f(q(t))dtds\|^2\\
    \leq& 2\|up\|^2+u^2\int_0^{u}\int_0^s\|\nabla f(q(t))\|^2dtds
\end{align*}
Inserting result from Lemma~\ref{lem:PosB} that $\|\nabla f(q(t))\|^2\leq 2M^2(\|q-\theta^*\|^2+t^2\|p\|^2)$\footnote{One needs to be careful that $p$ could be the updated momentum of discrete HMC and may not have normality, losing a lot of nicely properties. Nevertheless, Lemma~\ref{lem:PosB} holds for non-normally distributed initial momentum.} into the last inequality, we have
\[
\| q(u) - q \|^2\leq (2u^2+\frac{M^2u^6}{6})\|p\|^2+M^2u^4\|q-\theta^*\|^2
\]
Since $u\leq \eta \leq 1/(2\sqrt{M})$, we also have
\[
\| q(u) - q \|^2\leq \frac{24}{10}u^2\|p\|^2+M^2u^4\|q-\theta^*\|^2
\]
and inserting back to (\ref{pflem:PosDivB:1}) yields the claim.

\end{proof}

\begin{restatable}[Approximation error on momentum of classical HMC]{lemma}{lemMomDivB}\label{lem:MomDivB} Suppose $q$ is the associated initial position of momentum $p$.
Under Assumption \ref{assum:smooth}   and $\eta \leq 1/(2\sqrt{M})$,
\[
\|\He(p)-\barHe(p)\|^2\leq \frac{5}{2}\eta^4M^2\|p\|^2+\frac{2}{3}\eta^6M^4\|q-\theta^*\|^2%
\]
\end{restatable}
\begin{proof}

By definitions,
\[
\He(p)-\barHe(p)=p-\frac{\eta}{2}(\nabla f(q)+\nabla f(\He(q)))-(p-\int_0^\eta \nabla f(q(s))ds)
\]
Inserting  $\nabla f(q(s))=\nabla f(q)+\int_0^s \nabla^2 f(q(u)){p(u)}du$
into above, we get
\[
\He(p)-\barHe(p)=\frac{\eta}{2}(\nabla f(q)-\nabla f(\He(q)))+\int_0^\eta \int_0^s\nabla^2 f(q(u))p(u)duds
\]
Applying Lemma~\ref{lem:convex}, we can get
\begin{align*}
    \|\He(p)-\barHe(p)\|^2\leq & 2\|\frac{\eta}{2}(\nabla f(q)-\nabla f(\He(q)))\|^2+2\|\int_0^\eta \int_0^s\nabla^2 f(q(u))p(u)duds\|^2 \\
    \leq &2\|\frac{\eta}{2}(\nabla f(q)-\nabla f(\He(q)))\|^2+2\frac{\eta^2}{2}\int_0^\eta \int_0^s\|\nabla^2 f(q(u))p(u)\|^2duds\\
    =& \frac{\eta^2}{2}\|\nabla f(q)-\nabla f(\He(q))\|^2+\eta^2\int_0^\eta \int_0^s\|\nabla^2 f(q(u))p(u)\|^2duds.
\end{align*}
Since $f$ satisfies $M$-smoothness assumption, we further have

\begin{equation}\label{pflem:MomDivB:1}
\|\He(p)-\barHe(p)\|^2\leq \frac{\eta^2M^2}{2}\|q-\He(q)\|^2+\eta^2M^2\int_0^\eta \int_0^s\|p(u)\|^2duds
\end{equation}
To bound the RHS, setting $F=f$ in Lemma~\ref{lem:LocB} gives
\[
\|q-\He(q)\|^2\leq 2\eta^2 \|p\|^2+\eta^4 M^2\|q-\theta^*\|^2
\]
and at the same time,
{
\begin{align*}
    \|p(u)\|^2 =&\|p-\int_0^u \nabla f(q(s))ds\|^2
    \leq 2\|p\|^2+2\|\int_0^u \nabla f(q(s))ds\|^2\\
    \leq &2\|p\|^2+2u\int_0^u\| \nabla f(q(s))\|^2ds
    \leq 2\|p\|^2+2u\int_0^u2M^2(\|q-\theta^*\|^2+s^2\|p\|^2)ds\\
    = &(2+\frac{u^4M^2}{3})\|p\|^2+4u^2M^2\|q-\theta^*\|^2
    \leq \ 3\|p\|^2+4u^2M^2\|q-\theta^*\|^2,
\end{align*}
where the first inequality is by Cauthy-Schwarz inequality, the second inequality is by Lemma \ref{lem:convex}, the third inequality is by Lemma~\ref{lem:PosB}, and the last inequality is by $u\leq\eta\leq 1/2\sqrt{M}$.
}

Inserting these back to (\ref{pflem:MomDivB:1}) yields
\[
\|\He(p)-\barHe(p)\|^2\leq \frac{5}{2}\eta^4M^2\|p\|^2+\frac{2}{3}\eta^6M^4\|q-\theta^*\|^2.
\]
This finishes the proof.
\end{proof}
\subsection{Lemmas}
\begin{restatable}[  ]{lemma}{lemPosB} \label{lem:PosB}  Suppose $p$ is the associated initial momentum of position $q$.
Under Assumption \ref{assum:smooth}
\[
\|\nabla f(q)\|^2\leq M^2\|q-\theta^*\|^2,\qquad\|\nabla f(q(t))\|^2\leq 2M^2(\|q-\theta^*\|^2+t^2\|p\|^2)
\]
\end{restatable}

\begin{proof}
 Noting that $\|\nabla f(q(t))\|^2=\|\nabla f(q(t))-\nabla f(\theta^*)\|^2\leq M^2\|q(t)-\theta^*\|^2$, and by Lemma \ref{lem:oraXiP}
\[
\|q(t)-\theta^*\|\leq \|q-\theta^*\|+t\|p-0\|.
\]
Noting that $f$ satisfies $M$-smoothness assumption
\[
\|\nabla f(q(t))\|^2\leq M^2\|q(t)-\theta^*\|^2\leq  M^2(\|q-\theta^*\|+t\|p\|)^2.
\]
We finish the proof by applying Cauchy–Schwarz inequality on the RHS.
\end{proof}

\begin{restatable}[  ]{lemma}{lemdiffB} \label{lem:diffB} For any $q,p,q^\pi\in \mathbb{R}^d$, let $q^{(c)}_k=\widetilde{h}_{{\rm LF}}(f^{(c)},q,\nabla f (q),p_0,\eta,k)$, $ q_k=\sum_{c=1}^Nw_cq^{(c)}_k$ and $ q^\pi_k=\barHe^k(q^\pi)$. Under the smoothness Assumption \ref{assum:smooth}, we have
\[
\sum_{c=1}^Nw_c\mathbb{E}\|q^{(c)}_k-q_k\|^2\leq 2k^3\eta^4 \sum_{i=1}^{k-1}(16M^2\mathbb{E}\|q^\pi_i-q_i\|^2+16M^2\mathbb{E}\|q^\pi-q\|^2+4(k\eta)^2H)
\]
where $H:=\max_{i\leq K}\sum_{c=1}^Nw_c\mathbb{E}\|\frac{1}{i\eta}(\nabla f^{(c)}(q^\pi_i)-\nabla f^{(c)}(q^\pi)+\nabla f(q^\pi)-\nabla f(q^\pi_i))\|^2$ is a quantity measuring  the federated
 heterogeneity.
\end{restatable}
\begin{proof}

To prove this lemma, we use mathematical induction to remove the discrete heterogeneous term $\sum_{c=1}^Nw_c\mathbb{E}\|q^{(c)}_k-q_k\|^2$. In specific, we are going to use induction to prove that there are constants $C_1,C_2$ (to be determined later) such that
\begin{equation}\label{pflemdiffB:1}
 \sum_{c=1}^Nw_c\mathbb{E}\|q^{(c)}_j-q_j\|^2\leq (j-1)^3\eta^4 \sum_{k=1}^{j-1}(C_1M^2\mathbb{E}\|q^\pi_k-q_k\|^2+C_1M^2\mathbb{E}\|q^\pi-q\|^2+C_2(k\eta)^2H).
\end{equation}
First, for $j=1$, noting that $q^{(c)}_1=q+p-\frac{\eta}{2}(\nabla f^{(c)}(q)-\nabla f^{(c)}(q)+\nabla f(q))$ is constant for all $c$, thus
\[
\sum_{c=1}^Nw_c\mathbb{E}\|q^{(c)}_j-q_j\|^2=0, \qquad \text{when $j=1$.}
\]
Now, assume that (\ref{pflemdiffB:1}) holds for $j\leq k$. Next, we check it also holds for $k+1$ based on the assumption. By Lemma~\ref{lem:diffB:sublem}, we have
\begin{align*}
     &\sum_{c=1}^Nw_c\mathbb{E}\|q^{(c)}_{k+1}-q_{k+1}\|^2\\
     \leq &2(k+1)^3\eta^4 \sum_{i=1}^k\Bigl(4M^2\sum_{c=1}^Nw_c\mathbb{E}\|q^{(c)}_i-q_i\|^2+4M^2\mathbb{E}\|q^\pi_i-q_i\|^2+4M^2\mathbb{E}\|q^\pi-q\|^2+(i\eta)^2H\Bigr)\\
     \leq & 2(k+1)^3\eta^4 \sum_{i=1}^k\Bigl(4M^2i^3\eta^4\sum_{j=1}^{i-1}(C_1M^2\mathbb{E}\|q^\pi_j-q_j\|^2+C_1M^2\mathbb{E}\|q^\pi-q\|^2+C_2(j\eta)^2H)\Bigr)\\
     &+ 2(k+1)^3\eta^4  \sum_{i=1}^k\Bigl(4M^2\mathbb{E}\|q^\pi_i-q_i\|^2+4M^2\mathbb{E}\|q^\pi-q\|^2+(i\eta)^2H\Bigr)
\end{align*}
where the second inequality is by our induction assumption for $i\leq k$.

Noting that for any sequence $\{a_j\}$,
\[
\sum_{i=1}^ki^3\sum_{j=1}^{i-1}a_j=\sum_{j=1}^{k-1}\sum_{i=j+1}^ki^3a_j\leq \sum_{j=1}^{k-1}k^4a_j
\]
Therefore, we further have
\begin{align*}
     \sum_{c=1}^Nw_c\mathbb{E}\|q^{(c)}_{k+1}-q_{k+1}\|^2\leq &(k+1)^3\eta^4 \sum_{j=1}^k\Bigl((8M^4C_1k^4\eta^4+8M^2)\mathbb{E}\|q^\pi_j-q_j\|^2\\
     &+(8M^4C_1k^4\eta^4+8M^2)\mathbb{E}\|q^\pi-q\|^2+(8M^2C_2k^4\eta^4+2)H\Bigr)
\end{align*}
By choosing $C_1,C_2$ such that
\[
8M^4C_1k^4\eta^4+8M^2\leq C_1M^2,\qquad 8M^2C_2k^4\eta^4+2\leq C_2,
\]
we can finish the induction. Recall that $K\eta $ is set to be $\leq 1/(2\sqrt{M})$. One choice is that $C_1=16,C_2=4$. This finishes the proof.
\end{proof}

\begin{restatable}[  ]{lemma}{lemdiffB:sublem} \label{lem:diffB:sublem} Under the same notations and conditions of that of Lemma~\ref{lem:diffB},
\[
 \sum_{c=1}^Nw_c\mathbb{E}\|q^{(c)}_j-q_j\|^2\leq 2j^3\eta^4 \sum_{k=1}^{j-1}\Bigl(4M^2\sum_{c=1}^Nw_c\mathbb{E}\|q^{(c)}_k-q_k\|^2+4M^2\mathbb{E}\|q^\pi_k-q_k\|^2+4M^2\mathbb{E}\|q^\pi-q\|^2+(k\eta)^2H\Bigr)
\]
\end{restatable}

\begin{proof}
By definitions,
\begin{align*}
    q^{(c)}_j-q_j=&q+j\eta p_0-\frac{(j\eta)^2}{2}\nabla f(q)-\eta^2\sum_{k=1}^{j-1}\sum_{i=1}^k(\nabla f^{(c)}(q^{(c)}_i)-\nabla f^{(c)}(q))\\
    &-\sum_{c=1}^Nw_c\Bigl(q+j\eta p_0-\frac{(j\eta)^2}{2}\nabla f(q)-\eta^2\sum_{k=1}^{j-1}\sum_{i=1}^k(\nabla f^{(c)}(q^{(c)}_i)-\nabla f^{(c)}(q))\Bigr)\\
    =& \eta^2 \sum_{k=1}^{j-1}\sum_{i=1}^k\Bigl(\sum_{\widetilde{c}=1}^Nw_{\widetilde{c}}(\nabla f^{(\widetilde{c})}(q^{(\widetilde{c})}_i)-\nabla f^{(\widetilde{c})}(q))-\nabla f^{(c)}(q^{(c)}_i)+\nabla f^{(c)}(q)\Bigr).
\end{align*}
By Lemma~\ref{lem:convex}, we have
\begin{align*}
  \|q^{(c)}_j-q_j\|^2\leq &\frac{j(j-1)}{2} \sum_{k=1}^{j-1}\sum_{i=1}^k\| \eta^2\sum_{\widetilde{c}=1}^Nw_{\widetilde{c}}(\nabla f^{(\widetilde{c})}(q^{(\widetilde{c})}_i)-\nabla f^{(\widetilde{c})}(q))-\nabla f^{(c)}(q^{(c)}_i)+\nabla f^{(c)}(q)\|^2\\
=&  \frac{j(j-1)}{2} \sum_{i=1}^{j-1}\sum_{k=i}^{j-1}\eta^4\| \sum_{\widetilde{c}=1}^Nw_{\widetilde{c}}(\nabla f^{(\widetilde{c})}(q^{(\widetilde{c})}_i)-\nabla f^{(\widetilde{c})}(q))-\nabla f^{(c)}(q^{(c)}_i)+\nabla f^{(c)}(q)\|^2
\end{align*}
It follows that
\begin{equation}\label{lem:diffB:sublem:1}
    \|q^{(c)}_j-q_j\|^2\leq \frac{j^3\eta^4}{2} \sum_{i=1}^{j-1}\| \sum_{\widetilde{c}=1}^Nw_{\widetilde{c}}(\nabla f^{(\widetilde{c})}(q^{(\widetilde{c})}_i)-\nabla f^{(\widetilde{c})}(q))-\nabla f^{(c)}(q^{(c)}_i)+\nabla f^{(c)}(q)\|^2.
\end{equation}
The error $\sum_{\widetilde{c}=1}^Nw_{\widetilde{c}}(\nabla f^{(\widetilde{c})}(q^{(\widetilde{c})}_i)-\nabla f^{(\widetilde{c})}(q))-\nabla f^{(c)}(q^{(c)}_i)+\nabla f^{(c)}(q))$ can be split into several part as follow
\begin{align*}
    &\sum_{\widetilde{c}=1}^Nw_{\widetilde{c}}(\nabla f^{(\widetilde{c})}(q^{(\widetilde{c})}_i)-\nabla f^{(\widetilde{c})}(q))-\nabla f^{(c)}(q^{(c)}_i)+\nabla f^{(c)}(q)\\
    =& \underbrace{\sum_{\widetilde{c}=1}^Nw_{\widetilde{c}}(\nabla f^{(\widetilde{c})}(q^{(\widetilde{c})}_i)-\nabla f^{(\widetilde{c})}(q_i))-\nabla f^{( c)}(q^{(c)}_i)+\nabla f^{(c)}(q_i)}_{\text{discrete heterogeneity}}\\
    &+\underbrace{\nabla f^{(c)}(q^\pi_i)-\nabla f^{(c)}(q_i)+\nabla f(q_i)-\nabla f(q^\pi_i)}_{\text{convergence error}}\\
    &+\underbrace{\nabla f^{(c)}(q)-\nabla f^{(c)}(q^\pi)+\nabla f(q^\pi)-\nabla f(q)}_{\text{convergence error}}\\
    &+\underbrace{\nabla f^{(c)}(q^\pi)-\nabla f^{(c)}(q^\pi_i)+\nabla f(q^\pi_i)-\nabla f(q^\pi)}_{\text{oracle heterogeneity}}
\end{align*}

Again, by Lemma~\ref{lem:convex},
\begin{equation}\label{lem:diffB:sublem:2}
   \|\sum_{\widetilde{c}=1}^Nw_{\widetilde{c}}(\nabla f^{(\widetilde{c})}(q^{(\widetilde{c})}_i)-\nabla f^{(\widetilde{c})}(q))-\nabla f^{(c)}(q^{(c)}_i)+\nabla f^{(c)}(q)\|^2\leq 4(A^{(c)}+B^{(c)}+C^{(c)}+D^{(c)})
\end{equation}
where
\begin{align*}
 A^{(c)}:& \|\sum_{\widetilde{c}=1}^Nw_{\widetilde{c}}(\nabla f^{(\widetilde{c})}(q^{(\widetilde{c})}_i)-\nabla f^{(\widetilde{c})}(q_i))-\nabla f^{(\widetilde{c})}(q^{(c)}_i)+\nabla f^{(c)}(q_i)\|^2,\\
 B^{(c)}:&\|\nabla f^{(c)}(q^\pi_i)-\nabla f^{(c)}(q_i)+\nabla f(q_i)-\nabla f(q^\pi)\|^2\\
    C^{(c)}:&\|\nabla f^{(c)}(q)-\nabla f^{(c)}(q^\pi)+\nabla f(q^\pi)-\nabla f(q)\|^2,\\D^{(c)}:&\|\nabla f^{(c)}(q^\pi)-\nabla f^{(c)}(q^\pi_i)+\nabla f(q^\pi_i)-\nabla f(q^\pi)\|^2.
\end{align*}
Consider $A^{(c)}$ first, observe that
\[
\begin{split}
&\sum_{\widetilde{c}=1}^Nw_{\widetilde{c}}(\nabla f^{(\widetilde{c})}(q^{(\widetilde{c})}_i)-\nabla f^{(\widetilde{c})}(q_i))-\nabla f^{(\widetilde{c})}(q^{(c)}_i)+\nabla f^{(c)}(q_i)\\
=&\sum_{\widetilde{c}=1}^N\frac{w_{\widetilde{c}}-\mathbb{I}(\widetilde{c}=c)}{2(1-w_c)}\cdot 2(1-w_c)\Bigl(\nabla f^{(\widetilde{c})}(q^{(c)}_k)-\nabla f^{(\widetilde{c})}(q_k)\Bigr)
\end{split}
\]
and that $\sum_{\widetilde{c}=1}^N\frac{{ |w_{\widetilde{c}}-\mathbb{I}(\widetilde{c}=c)|}}{2(1-w_c)}=1$, by Lemma~\ref{lem:convex}

\begin{align*}
    \sum_{c=1}^Nw_cA^{(c)}\leq& \sum_{c=1}^Nw_c\sum_{\widetilde{c}=1}^N|w_c-\mathbb{I}(\widetilde{c}=c)|2(1-w_c)\|\nabla f^{(\widetilde{c})}(q^{(c)}_k)-\nabla f^{(\widetilde{c})}(q_k)\|^2\\
=&2\sum_{\widetilde{c}=1}^N w_{\widetilde{c}}( \underbrace{\sum_{c\neq \widetilde{c}}^Nw_c(1-w_c)}_{:\leq \sum_{c=1}^Nw_c=1}+(1-w_{\widetilde{c}})^2)\|\nabla f^{(\widetilde{c})}(q^{(c)}_k)-\nabla f^{(\widetilde{c})}(q_k)\|^2   \\
\leq &4\sum_{\widetilde{c}=1}^N w_{\widetilde{c}}\|\nabla f^{(\widetilde{c})}(q^{(c)}_k)-\nabla f^{(\widetilde{c})}(q_k)\|^2
\end{align*}
Further, by smoothness assumption,
\begin{equation}\label{lem:diffB:sublem:3}
    \sum_{c=1}^Nw_cA^{(c)}\leq 4\sum_{c=1}^N w_{c}M^2\|q^{(c)}_k-q_k\|^2
\end{equation}
Similarly, for $B^{(c)}$ and $C^{(c)}$, we have
\begin{align}
    \begin{split}\label{lem:diffB:sublem:4}
        \sum_{c=1}^Nw_cB^{(c)}\leq &4\sum_{c=1}^N w_cM^2\|q^\pi_k-q_k\|^2=4M^2\|q^\pi_k-q_k\|^2 \\
        \sum_{c=1}^Nw_cC^{(c)}\leq &4\sum_{c=1}^N w_cM^2\|q^\pi-q\|^2=4M^2\|q^\pi-q\|^2
    \end{split}
\end{align}
For $D^{(c)}$, by definitions, we directly have
\begin{equation}\label{lem:diffB:sublem:5}
    \sum_{c=1}^Nw_c\mathbb{E}D^{(c)}\leq (k\eta)^2H
\end{equation}
Combining (\ref{lem:diffB:sublem:1})-(\ref{lem:diffB:sublem:5}), we have
\[
 \sum_{c=1}^Nw_c\mathbb{E}\|q^{(c)}_j-q_j\|^2\leq 2j^3\eta^4 \sum_{k=1}^{j-1}\Bigl(4M^2\sum_{c=1}^Nw_c\mathbb{E}\|q^{(c)}_k-q_k\|^2+4M^2\mathbb{E}\|q^\pi_k-q_k\|^2+4M^2\mathbb{E}\|q^\pi-q\|^2+(k\eta)^2H\Bigr)
\]
This finishes the proof.
\end{proof}

\begin{restatable}[$0$-th order Approximation Bound]{lemma}{lemLocB}\label{lem:LocB} Let $q_0,p_0\in \mathbb{R}^d$ be two vectors. For any $F(\cdot):\mathbb{R}^d \mapsto\mathbb{R}$ satisfying smoothness Assumption \ref{assum:smooth}, then $q_K=\widetilde{h}_{\rm LF}(F,q_0,\nabla f(q_0),p_0,\eta,k)$ satisfies
\begin{align}
       & \| q_k-q_0\|^2\leq 2(k\eta)^2\|p_0\|^2+(k\eta)^4M^2\|q_0-\theta^*\|^2\label{eqn:LocB:a1}\\
      & \| \nabla F (q_k)-\nabla F(q_0)\|^2\leq 2M^2(k\eta)^2\|p_0\|^2+(k\eta)^4M^4\|q_0-\theta^*\|^2\label{eqn:LocB:a2}
\end{align}
\end{restatable}
Note that we actually prove \ref{pflem:LocB:1}, which is an interesting result in the sense that the $k$-th leapfrog step can be bounded by the shift vector $\nabla f(q_0)$ rather than $\nabla F(q_0)$.

\begin{proof}
We are going to prove
\begin{equation}\label{pflem:LocB:1}
    \| q_k-q_0\|^2\leq 2(k\eta)^2\|p_0\|^2+(k\eta)^4\|\nabla f(q_0)\|^2.
\end{equation}

We prove claim \eqref{pflem:LocB:1} by induction.
Then the claim \eqref{eqn:LocB:a1}  and \eqref{eqn:LocB:a2} follows directly from \eqref{pflem:LocB:1} and $M$-smoothness Assumption \ref{assum:smooth} applied with $\nabla f(\theta^*)=0$.

First, according to the definitions and Algorithm~\ref{alg:surHMC}, we can write the full expression of $q_k$ as follow
\[
   q_k=q_0+k\eta p_0-\frac{(k\eta)^2}{2}\nabla f(q_0)-\eta^2\sum_{j=1}^{k-1}(k-j) (\nabla F(q_j)-\nabla F(q_0))
\]
which is proved by Lemma~\ref{lem:eqexp}.

For claim \eqref{pflem:LocB:1} with $k=1$, by  Cauchy–Schwarz inequality, we have that
\[
 \|q_1-q_0\|^2
    =\|\eta p_0-\frac{\eta^2}{2}\nabla f(q_0)\|^2
    \leq  2\|\eta p_0\|^2+2\|\frac{\eta^2}{2}\nabla f(q_0)\|^2
\]

Next, we assume that the claim \eqref{pflem:LocB:1} holds for $j=1,2,\ldots,k-1$.
By direct calculations, for any $c_1,c_2,c_3>0$ such that $\sum_ic_i=1$  we have
\begin{align*}
   &\|q_k-q_0\|^2\\
        \leq &   \frac{1}{c_1}(k\eta)^2\| p_0\|^2 +\frac{(k\eta)^4}{4c_2}\|\nabla f(q_0)\|^2+\sum_{j=1}^{k-1}\frac{(k-j)\eta^4k^3}{6j c_3}\|\nabla F(q_j)-\nabla F(q_0)\|^2\\
    \leq &  \frac{1}{c_1} (k\eta)^2\| p_0\|^2 +\frac{(k\eta)^4}{4c_2}\|\nabla f(q_0)\|^2+\sum_{j=1}^{k-1}\frac{(k-j)\eta^4k^3M^2}{6j c_3}\|q_j-q_0\|^2\\
    \leq &   \frac{1}{c_1} (k\eta)^2\| p_0\|^2 +\frac{(k\eta)^4}{4c_2}\|\nabla f(q_0)\|^2+\sum_{j=1}^{k-1}\frac{(k-j)\eta^4k^3M^2}{6j c_3}
    \cdot (2(j\eta)^2\|p_0\|^2+(j\eta)^4\|\nabla f(q_0)\|^2)\\
    \leq &   (\frac{1}{c_1}+\frac{M^2}{6 c_3}\frac{2(k\eta)^4}{6})(k\eta)^2\| p_0\|^2+(\frac{1}{4c_2}+\frac{M^2}{6 c_3}\frac{(k\eta)^4}{20})(k\eta)^4\|\nabla f(q_0)\|^2
\end{align*}
where the first inequality is due to Lemma \ref{lem:convex} and noting that  $c_1+c_2+\frac{\sum_{j=1}^{k-1}(k-j)j}{(k^3-k)/6}c_3=1$;  the second inequality is due to smoothness Assumption \ref{assum:smooth}; the third inequality holds by the induction assumption; the fourth inequality is by $\sum_{j=1}^{k-1}(k-j)j=\frac{k^3-k}{6}\leq \frac{k^3}{6}$ and  $\sum_{j=1}^{k-1}(k-j)j^3=\frac{k^5}{20}-\frac{k^3}{12}+\frac{1}{20}k\leq \frac{k^5}{20}$. The completion of the induction follows by step size assumption $K\eta\leq 1/\sqrt{M}$ and setting $(c_1,c_2,c_3)=(\frac{3}{5},\frac{1}{3},\frac{1}{15})$ in the last inequality. Therefore, we conclude the proof of \eqref{pflem:LocB:1}.
\end{proof}

\begin{restatable}[]{lemma}{lemeqexp}\label{lem:eqexp} For $q_k=\widetilde{h}_{\rm LF}(F,q_0,\nabla f(q_0),p_0,\eta,k)$ and $p_k=p_{k-1}-\eta(\nabla F(q_{k-1})+\nabla F(q_k) )/2+\eta (\nabla F(q_0)-\nabla f(q_0))$, we can rewrite as
\begin{align}
       & q_k=q_0+k\eta p_0-\frac{(k\eta)^2}{2}\nabla f(q_0)-\eta^2\sum_{i=1}^{k-1}(k-i) (\nabla F(q_j)-\nabla F(q_0))\label{lem:eqexp:1}\\
   &p_k=p_0-k\eta\nabla f(q_0)-\eta\sum_{j=1}^{k-1}(\nabla F(q_j)-\nabla F(q_0))-\frac{\eta}{2}(\nabla F(q_k)-\nabla F(q_0))\label{lem:eqexp:2}.
\end{align}
\end{restatable}

\begin{proof}
For simplicity of notations, we use $\widetilde{f}(\cdot):=F(\cdot)-F(q_0)+f(q_0)$. According to Algorithm~\ref{alg:surHMC},
\begin{equation}\label{pflem:eqexp:1}
q_k=q_{k-1}+\eta p_{k-1}-\frac{\eta^2}{2}\nabla \widetilde{f}(q_{k-1}),\qquad p_k=p_{k-1}-\eta(\nabla \widetilde{f}(q_{k-1})+\nabla \widetilde{f}(q_k) )/2
\end{equation}
Summing from $k=1,\ldots,k$ on both side of the second equation of (\ref{pflem:eqexp:1}), we can get
\[
p_k=p_0-\frac{\eta}{2}\nabla \widetilde{f}(q_0)-\eta\sum_{j=1}^{k-1}\nabla \widetilde{f}(q_j)-\frac{\eta}{2}\nabla \widetilde{f}(q_k).
\]
This proves~(\ref{lem:eqexp:2}). Inserting this into the first equality of (\ref{pflem:eqexp:1}), we can get
\[
q_k=q_{k-1}+p_0-\frac{\eta^2}{2}\nabla \widetilde{f}(q_0)-\eta^2\sum_{j=1}^{k-1}\nabla \widetilde{f}(q_j)
\]
Summing from $k=1,\ldots,k$ on both side yields
\[
q_k=q_0+k\eta p_0-\frac{(k\eta)^2}{2}\nabla f(q_0)-\eta^2\sum_{i=1}^{k-1}(k-i) (\nabla F(q_j)-\nabla F(q_0))
\]
which proves~(\ref{lem:eqexp:1}).
\end{proof}

\begin{restatable}[  ]{lemma}{lemAveLocB} \label{lem:AveLocB}
Under the same notations of Theorem~\ref{thm:AFLHMC} and Assumption \ref{assum:smooth}, for $K\eta \leq 1/(2\sqrt{M})$, we have
\begin{align*}
  &\|p_{t,k}\|^2\leq 5\|p_t\|^2+5M^2(k\eta)^2\|\theta_t-\theta^*\|^2\\
  &\|\theta_{t,k}-\theta^*\|^2\leq 4(k\eta)^2\|p_t\|^2+3\|\theta_t-\theta^*\|^2.
\end{align*}
\end{restatable}

\begin{proof}
Recall that $p_{t,k}:=\sum_{c=1}^Nw_cp^{(c)}_{t,k}$, and note that by Lemma~\ref{lem:MomB} that $\|p^{(c)}_{t,k}\|^2\leq 5\|p_t\|^2+5M^2(k\eta)^2\|\theta_t-\theta^*\|^2$, we have
\[
\|p_{t,k}\|^2\leq \sum_{c=1}^Nw_c\|p^{(c)}_{t,k}\|^2\leq 5\|p_t\|^2+5M^2(k\eta)^2\|\theta_t-\theta^*\|^2.
\]
This proves the first claim.

For the second claim, first, by convexity of $\|\cdot\|^2$ and Lemma~\ref{lem:convex}
\[
\|\theta_{t,k}-\theta^*\|^2 \leq 2\|\theta_{t,k}-\theta_t\|^2+2\|\theta_t-\theta^*\|^2
 \leq \sum_{c=1}^Nw_c 2\|\theta^{(c)}_{t,k}-\theta_t\|^2 +2\|\theta_t-\theta^*\|^2.
\]
At the same time, for $\|\theta^{(c)}_{t,k}-\theta_t\|^2 $,  Lemma~\ref{lem:LocB} states that $\|\theta^{(c)}_{t,k}-\theta_t\|^2 \leq 2(k\eta)^2\|p_t\|^2+(k\eta)^4M^2\|\theta_t-\theta^*\|^2$. And so
\[
\|\theta_{t,k}-\theta^*\|^2
 \leq 4(k\eta)^2\|p_t\|^2+(2+2(k\eta)^4M^2)\|\theta_t-\theta^*\|^2\leq 4(k\eta)^2\|p_t\|^2+3\|\theta_t-\theta^*\|^2
\]
where the last inequality is due to $k\eta\leq 1/(2\sqrt{M})$. This finishes the proof.
\end{proof}

\begin{restatable}[  ]{lemma}{lemMomB} \label{lem:MomB} Suppose $q_k=\widetilde{h}_{\rm LF}(F,q_0,\nabla f(q_0),p_0,\eta,k)$ and $p_k=p_{k-1}-\eta(\nabla F(q_{k-1})+\nabla F(q_k) )/2+\eta (\nabla F(q_0)-\nabla f(q_0))$.
Under Assumption \ref{assum:smooth}, for $K\eta \leq 1/(2\sqrt{M})$, we have
\[
\|p_j\|^2\leq 5\|p_0\|^2+5M^2(j\eta)^2\|q_0-\theta^*\|^2.
\]
\end{restatable}

\begin{proof}

By definitions and Algorithm~\ref{alg:surHMC}, we have
\[
p_k=p_0-k\eta\nabla f(q_0)-\eta\sum_{j=1}^{k-1}(\nabla F(q_j)-\nabla F(q_0))-\frac{\eta}{2}(\nabla F(q_k)-\nabla F(q_0))
\]
By   Cauchy–Schwarz inequality and $M$-smoothness assumption,
\begin{align*}
    \|p_k\|^2\leq &4\|p_0\|^2+4\|k\eta\nabla f(q_0)\|^2+4\|\eta\sum_{j=1}^{k-1}(\nabla F(q_j)-\nabla F(q_0))\|^2+4\|\frac{\eta}{2}(\nabla F(q_k)-\nabla F(q_0))\|^2\\
    \leq & 4\|p_0\|^2+4\|k\eta\nabla f(q_0)\|^2+4(k-1)\sum_{j=1}^{k-1}\eta^2\|\nabla F(q_j)-\nabla F(q_0)\|^2+4\|\frac{\eta}{2}(\nabla F(q_k)-\nabla F(q_0))\|^2\\
    \leq & 4\|p_0\|^2+4(k\eta)^2\|\nabla f(q_0)\|^2+4(k-1)\sum_{j=1}^{k-1}\eta^2M^2\|q_j-q_0\|^2+\eta^2M^2\|q_k-q_0\|^2\\
    \leq &4\|p_0\|^2+4(k\eta)^2\|\nabla f(q_0)\|^2+4(k\eta)^2M^2\max_{j\leq k}\|q_j-q_0\|^2
\end{align*}
where the first two inequalities are by Lemma \ref{lem:convex} and the third inequality is by $M$-smoothness assumption.

Note that Lemma~\ref{lem:LocB} gives that $\|q_j-q_0\|^2\leq 2(j\eta)^2\|p_0\|^2+(j\eta)^4M^2\|q_0-\theta^*\|^2$.  Inserting this into the above inequality yields
\[
\|p_k\|^2\leq 4(1+2(k\eta)^4M^2)\|p_0\|^2+4(k\eta)^2(1+M^2(k\eta)^4)M^2\|q_0-\theta^*\|^2.
\]
The second claim follows by noting that $k\eta \leq 1/(2\sqrt{M})$.
\end{proof}
\begin{restatable}[Jensen's inequality]{lemma}{lemconvex}\label{lem:convex}
For any vectors $\{x_i\}_{i=1}^n$ and positive constants $\{\lambda_i\}_{i=1}^n$ with $\sum\lambda_i=1$, due to the convexity of $\|\cdot\|^2$, we have
\[
\begin{split}
&\|\sum_{i}\lambda_ix_i\|^2\leq  \sum_i\lambda_i\|x_i\|^2;\\
&\|\sum_{i}x_i\|^2\leq \sum_i\lambda_i\|x_i/\lambda_i\|^2=\sum_i \frac{1}{\lambda_i}\|x_i\|^2.
\end{split}
\]
For any vector-value function $x(t)$ and non-negative integrable function $\lambda(t)$,
\[
\begin{split}
&\|\int \lambda(t)x(t)dt\|^2\leq \int\lambda(t)dt\int\lambda(t)\|x(t)\|^2dt.\\
&\|\int \lambda(t)x(t)dt\|^2\leq \int\lambda(t)\|x(t)\|^2dt, \quad \mbox{ if } \int\lambda(t)dt=1.
\end{split}
\]
\end{restatable}

\section{\revise{Study on heterogeneity metric $H$}}\label{section:H}
\subsection{Bound on $H$}
Recall that $H:=\max_{i\leq K}\sum_{c=1}^Nw_c\mathbb{E}\|\frac{1}{i\eta}(\nabla f^{(c)}(\bar{{\cal H}}_{\eta}^i(\theta^\pi))-\nabla f^{(c)}(\theta^\pi)+\nabla f(\theta^\pi)-\nabla f(\bar{{\cal H}}_{\eta}^i(\theta^\pi)))\|^2$
By the $M$-Lipschitz condition,
\[
H \le 4 \max_{1 \le i \le K} \sum_{c=1}^{N} w_c \mathbb{E} \left\| \frac{M}{i\eta} \left( \bar{\mathcal{H}}^i_{\eta}(\theta^{\pi}) - \theta^{\pi} \right) \right\|^{2}.
\]
Moreover,
\[
\bar{\mathcal{H}}^i_{\eta}(\theta^{\pi}) - \theta^{\pi} = \int_{0}^{i\eta} p(t)\,dt.
\]
where $p(t)$ has a stationary marginal multivariate normal distribution. Overall, we can obtain that
\[
H \le C M^{2} d
\]
for some absolute constant $C$.
\subsection{Solutions of $H$ on Gaussian examples}
Give normal clients with same covariance structure (e.g., \(N(1, I)\) and \(N(-1, I)\), and the global is \(N(0,I)\)), then
\[
\nabla f^{(c)}(\theta_2) - \nabla f^{(c)}(\theta_1) + \nabla f(\theta_1) - \nabla f(\theta_2) \equiv 0,
\]
meaning that \(H \equiv 0\) always.

For non-equal covariance case, e.g., \(N(0, 3I/4)\) and \(N(0, 3I/2)\), and the global is \(N(0,I)\), the HMC trajectory for the global has a closed form, i.e., \(\theta_t = \theta_0 \cos(t) + p_0 \sin(t)\) elementwisely. The calculation of \(H\) eventually reduces to
\[
d \times \sup_t \mathbb{E}_{\theta,p\sim N(0,I^2)}\left|\frac{\theta(1-\cos(t))-p\sin(t)}{t}\right|^2\Big/9
= \frac{2d}{9}\times\sup_t\frac{1-\cos(t)}{t^2}
= \frac{d}{9}.
\]

\section{Experimental Details}
\subsection{Additional Details for Figure~\ref{fig:logreg}}
\label{app:fig1-grid}

Table~\ref{tab:fig1-grid} reports the stepsize and leapfrog grids used for
the simulated Bayesian logistic regression experiment in Figure~\ref{fig:logreg}.
The same grid is used for both panels. FA-LD uses $T=10$ and $K=1$ for every
stepsize. FA-HMC uses $T=10$ and the $K$ values listed in the table. AFHMC uses
the candidate leapfrog counts $K_{\mathrm{A}}\in
\mathcal{K}_{\mathrm{HMC}}(\eta)\cup
10\mathcal{K}_{\mathrm{HMC}}(\eta)$. For each $(\mathrm{algorithm},\eta)$
pair, Figure~\ref{fig:logreg} reports the best-performing $K$ from this grid.

\begin{table*}[t]
  \centering
  \small
  \caption{Candidate grid for the simulated Bayesian logistic regression
  experiment. Here $\mathcal{K}_{\mathrm{HMC}}(\eta)$ denotes the FA-HMC
  leapfrog grid at stepsize $\eta$; the AFHMC grid is
  $\mathcal{K}_{\mathrm{HMC}}(\eta)\cup 10\mathcal{K}_{\mathrm{HMC}}(\eta)$.}
  \label{tab:fig1-grid}
  \begin{tabular}{c c c c}
    \toprule
    Stepsize $\eta$ & FA-LD $(T,K)$ & FA-HMC $(T,\mathcal{K}_{\mathrm{HMC}})$
    & AFHMC $\mathcal{K}_{\mathrm{A}}$ \\
    \midrule
    $5\times 10^{-4}$ & $(10,\{1\})$ & $(10,\{500,2000\})$
      & $\{500,2000,5000,20000\}$ \\
    $10^{-3}$ & $(10,\{1\})$ & $(10,\{300,1000\})$
      & $\{300,1000,3000,10000\}$ \\
    $2\times 10^{-3}$ & $(10,\{1\})$ & $(10,\{150,500\})$
      & $\{150,500,1500,5000\}$ \\
    $5\times 10^{-3}$ & $(10,\{1\})$ & $(10,\{50,200\})$
      & $\{50,200,500,2000\}$ \\
    $10^{-2}$ & $(10,\{1\})$ & $(10,\{30,100\})$
      & $\{30,100,300,1000\}$ \\
    $2\times 10^{-2}$ & $(10,\{1\})$ & $(10,\{10,50\})$
      & $\{10,50,100,500\}$ \\
    $5\times 10^{-2}$ & $(10,\{1\})$ & $(10,\{2,5,10,20\})$
      & $\{2,5,10,20,50,100,200\}$ \\
    $10^{-1}$ & $(10,\{1\})$ & $(10,\{1,2,5,10\})$
      & $\{1,2,5,10,20,50,100\}$ \\
    $2\times 10^{-1}$ & $(10,\{1\})$ & $(10,\{1,2,5\})$
      & $\{1,2,5,10,20,50\}$ \\
    $5\times 10^{-1}$ & $(10,\{1\})$ & $(10,\{1,2,5\})$
      & $\{1,2,5,10,20,50\}$ \\
    $1$ & $(10,\{1\})$ & $(10,\{1,2\})$
      & $\{1,2,10,20\}$ \\
    \bottomrule
  \end{tabular}
\end{table*}

\subsection[Additional Details for the Gaussian Dimension-Scaling Experiments]{\revise{Additional Details for the Gaussian Dimension-Scaling Experiments}}
\label{app:gaussian-extra}

\revise{\textbf{Low-heterogeneity target.} For each dimension $d$ and each of the $N=10$ clients, we draw $10^5$ samples from $\mathcal{N}(16.2\mathbf{1}_d,1.6I_d)$ and set the local target of client $c$ to $\mathcal{N}(\hat\mu_c,\hat\sigma_c^2I_d)$, where $\hat\mu_c$ is the sample mean and $\hat\sigma_c^2$ is the coordinate-averaged sample variance. The global target is fitted in the same way to the samples pooled over all clients, and ${\cal W}_2^2$ is computed as in the heterogeneous experiment. The heterogeneity therefore only comes from sampling error: across all dimensions and seeds, the root-mean-square deviation of the client means from the global mean is between $3.6\times 10^{-3}$ and $4.5\times 10^{-3}$ per coordinate, compared with $11.1$ in the heterogeneous experiment, where the client means are $20$ and $1$ and the global mean is $16.2$. AFHMC runs one trajectory of $K(d)$ surrogate leapfrog steps per round and uses the stopping rule of Section~\ref{sec:numerical} (${\cal W}_2^2<0.1$ for $20$ consecutive checks, at most $1000$ rounds). The random seed controls both the client samples and the HMC noise. Table~\ref{tab:lowhet} reports the first-crossing rounds for every seed.}

\begin{table}[t]
\centering
\caption{\revise{First communication round at which AFHMC reaches ${\cal W}_2^2<0.1$ on the low-heterogeneity Gaussian target, with $K(d)=\lfloor 50(d/100)^{1/2}\rceil$ and $\eta(d)=0.006(d/100)^{-1/2}$.}}
\label{tab:lowhet}
\revise{\begin{tabular}{rrrcc}
\toprule
$d$ & $K(d)$ & $\eta(d)$ & Rounds (seeds 0, 1, 2) & Mean $\pm$ sd \\
\midrule
2 & 7 & 0.04243 & 161, 153, 149 & $154.33 \pm 6.11$ \\
50 & 35 & 0.008485 & 213, 212, 212 & $212.33 \pm 0.58$ \\
100 & 50 & 0.006 & 220, 220, 220 & $220.00 \pm 0.00$ \\
150 & 61 & 0.004899 & 229, 229, 229 & $229.00 \pm 0.00$ \\
200 & 71 & 0.004243 & 231, 231, 230 & $230.67 \pm 0.58$ \\
250 & 79 & 0.003795 & 237, 237, 237 & $237.00 \pm 0.00$ \\
300 & 87 & 0.003464 & 238, 238, 237 & $237.67 \pm 0.58$ \\
350 & 94 & 0.003207 & 240, 240, 240 & $240.00 \pm 0.00$ \\
400 & 100 & 0.003 & 245, 245, 245 & $245.00 \pm 0.00$ \\
450 & 106 & 0.002828 & 247, 248, 247 & $247.33 \pm 0.58$ \\
500 & 112 & 0.002683 & 248, 248, 248 & $248.00 \pm 0.00$ \\
\bottomrule
\end{tabular}}
\end{table}

\revise{\textbf{FA-HMC baselines at an equal number of server exchanges.} Theorem~4.1 of \cite{liang2025bayesian} shows that, under strong convexity, Hessian smoothness and bounded stochastic-gradient variance, FA-HMC attains ${\cal W}_2\leq\epsilon$ when
\[
\eta^2 = C\min\left\{\frac{1}{K^2L},\ \frac{\epsilon}{K^2\sqrt{d}\,T},\ \frac{\epsilon^2}{K^2dT^2(1-\rho)N},\ \frac{\epsilon^2}{Kd\sum_{c=1}^{N}w_c^2\sigma_g^2}\right\}.
\]
In our Gaussian experiment all clients share the momentum, so $\rho=1$, and the gradients are exact, so $\sigma_g^2=0$; the last two terms therefore impose no constraint. The remaining dimension-dependent term gives $\eta\propto (K\sqrt{T}d^{1/4})^{-1}$. With $T=10$ and $K_{\mathrm{F}}(d)=\max\{1,\lfloor K(d)/10\rceil\}\propto d^{1/3}$, this yields $\eta(d)\propto d^{-7/12}$, which we anchor at $d=100$ as $\eta(d)=0.006(d/100)^{-7/12}$. The second schedule is the dimension-scaling setting of \cite{liang2025bayesian}, $K_{\mathrm{F}}=5$, $T=10$ and $\eta(d)=0.02d^{-1/4}$. Both schedules are run for $2R(d)$ rounds, where $R(d)$ is the AFHMC first-crossing round, so that FA-HMC uses the same number of server exchanges as AFHMC. Table~\ref{tab:fahmc-equal} reports the resulting errors together with the original matched-budget setting. These schedules are theory-guided choices rather than an exhaustive search over $(\eta,K,T)$, and each setting is run once.}

\begin{table}[t]
\centering
\small
\setlength{\tabcolsep}{4pt}
\caption{\revise{FA-HMC ${\cal W}_2^2$ on the heterogeneous Gaussian target. $R(d)$ is the AFHMC first-crossing round for ${\cal W}_2^2<0.1$. The original setting uses the AFHMC stepsize for $R(d)$ rounds; the two FA-HMC-specific schedules are run for $2R(d)$ rounds. All settings use $T=10$; the original and Theorem-4.1 settings use $K_{\mathrm{F}}(d)$, and the second setting uses $K_{\mathrm{F}}=5$.}}
\label{tab:fahmc-equal}
\revise{\begin{tabular}{rrrrrrrrrr}
\toprule
 & & & & \multicolumn{2}{c}{Original ($R(d)$)} & \multicolumn{2}{c}{$K_{\mathrm{F}}{=}5$ ($2R(d)$)} & \multicolumn{2}{c}{Thm.~4.1 ($2R(d)$)} \\
\cmidrule(lr){5-6}\cmidrule(lr){7-8}\cmidrule(lr){9-10}
$d$ & $R(d)$ & $2R(d)$ & $K_{\mathrm{F}}(d)$ & $\eta$ & ${\cal W}_2^2$ & $\eta$ & ${\cal W}_2^2$ & $\eta$ & ${\cal W}_2^2$ \\
\midrule
2 & 42 & 84 & 1 & 0.04243 & 327.9 & 0.01682 & 13.0 & 0.05878 & 85.7 \\
50 & 162 & 324 & 4 & 0.008485 & 4092.3 & 0.007521 & 752.1 & 0.00899 & 960.2 \\
100 & 220 & 440 & 5 & 0.006 & 7621.3 & 0.006325 & 1682.8 & 0.006 & 2213.9 \\
150 & 263 & 526 & 6 & 0.004899 & 9523.9 & 0.005715 & 2694.0 & 0.004736 & 2774.3 \\
200 & 294 & 588 & 6 & 0.004243 & 15968.2 & 0.005318 & 3911.7 & 0.004005 & 6298.0 \\
250 & 320 & 640 & 7 & 0.003795 & 16013.5 & 0.00503 & 5236.5 & 0.003516 & 5827.7 \\
300 & 347 & 694 & 7 & 0.003464 & 22009.6 & 0.004806 & 6444.4 & 0.003161 & 9424.0 \\
350 & 368 & 736 & 8 & 0.003207 & 20227.2 & 0.004624 & 7867.1 & 0.002889 & 7877.5 \\
400 & 393 & 786 & 8 & 0.003 & 25525.1 & 0.004472 & 9014.2 & 0.002673 & 11122.8 \\
450 & 403 & 806 & 8 & 0.002828 & 32540.4 & 0.004342 & 11000.1 & 0.002495 & 15877.5 \\
500 & 430 & 860 & 8 & 0.002683 & 38053.9 & 0.004229 & 11872.9 & 0.002346 & 19755.7 \\
\bottomrule
\end{tabular}}
\end{table}

\subsection{Additional Details for MNIST Logistic Regression}
\label{app:fmnist-extra}
To create heterogeneous federated learning scenario, the FMNIST training set is partitioned across $N=10$ clients using a label-skewed Dirichlet split. Specifically, for each class $c$, we sample a client-proportion vector
\[
  p_c \sim \mathrm{Dirichlet}(\alpha \mathbf{1}_N),
  \qquad \alpha=0.1,
\]
and distribute the examples of class $c$ across clients according to $p_c$. This produces heterogeneous clients with different label distributions.
\begin{figure}[t]
  \centering
  \includegraphics[width=0.85\linewidth]{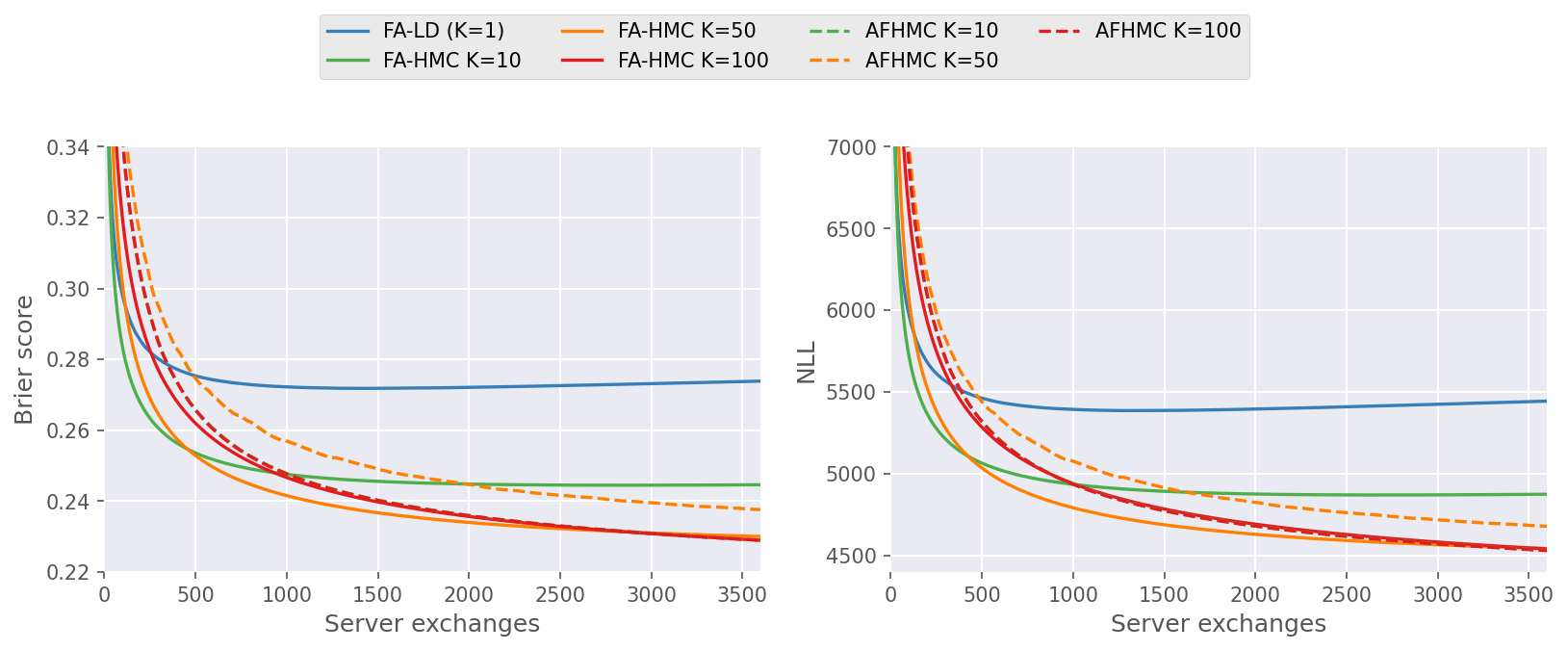}
  \caption{Additional predictive-quality metrics for Bayesian logistic
  regression on heterogeneous FMNIST \revise{against the number of server exchanges (one per round for FA-LD and FA-HMC, two per round for AFHMC)}. Left:
  Brier score. Right: negative log-likelihood. The data are partitioned across
  $N=10$ clients using a label-skewed Dirichlet split with concentration
  parameter $\alpha=0.1$.}
  \label{fig:fmnist-brier-nll}
\end{figure}

Each configuration is repeated over five random seeds. The reported values in Table~\ref{tab:fmnist-final-se} are the final mean and standard error (SE) across seeds at \revise{$3600$ server exchanges, i.e., at round $3600$ for FA-LD and FA-HMC and at round $1800$ for AFHMC}. Since the stepsize is tuned separately for each metric, the selected stepsize may differ between Accuracy, Brier score, and NLL for the same method and $K$. \revise{Table~\ref{tab:fmnist-trajectory} reports the accuracy at intermediate budgets. At $3600$ server exchanges, AFHMC with $K\in\{10,100\}$ attains the highest accuracy and the lowest NLL, and ties with FA-HMC ($K=100$) in Brier score; FA-HMC with $K=50$ has lower Brier score and NLL at intermediate budgets.}

\revise{\textbf{Computational cost.} Between two synchronizations, each client performs $50$ leapfrog steps in FA-LD and FA-HMC, i.e., $100$ mini-batch gradient evaluations per round, and $K$ leapfrog steps plus one gradient evaluation for the drift correction in AFHMC, i.e., $2K+1$ evaluations per round. At $3600$ server exchanges, AFHMC with $K=10$ therefore uses $1800\times 21=37{,}800$ gradient evaluations per client, compared with $3600\times 100=360{,}000$ for FA-HMC.}

\begin{table}[t]
\centering
\caption{Final performance on heterogeneous FMNIST at \revise{$3600$ server exchanges}. Entries are mean $\pm$ standard error over five random seeds.}
\label{tab:fmnist-final-se}
\revise{\begin{tabular}{llcccc}
\toprule
Metric & Method & $K$ & Rounds & $\eta$ & Mean $\pm$ SE \\
\midrule
Accuracy & FA-LD  & 1 & 3600 & $1{\times}10^{-4}$ & $0.8075 \pm 0.0031$ \\
 & FA-HMC & 10 & 3600 & $2{\times}10^{-5}$ & $0.8274 \pm 0.0036$ \\
 & FA-HMC & 50 & 3600 & $5{\times}10^{-6}$ & $0.8384 \pm 0.0007$ \\
 & FA-HMC & 100 & 3600 & $2{\times}10^{-6}$ & $0.8408 \pm 0.0003$ \\
 & AFHMC  & 10 & 1800 & $2{\times}10^{-5}$ & $0.8422 \pm 0.0005$ \\
 & AFHMC  & 50 & 1800 & $5{\times}10^{-6}$ & $0.8402 \pm 0.0005$ \\
 & AFHMC  & 100 & 1800 & $2{\times}10^{-6}$ & $0.8422 \pm 0.0004$ \\
\midrule
Brier & FA-LD  & 1 & 3600 & $1{\times}10^{-4}$ & $0.2739 \pm 0.0048$ \\
 & FA-HMC & 10 & 3600 & $2{\times}10^{-5}$ & $0.2447 \pm 0.0047$ \\
 & FA-HMC & 50 & 3600 & $5{\times}10^{-6}$ & $0.2300 \pm 0.0015$ \\
 & FA-HMC & 100 & 3600 & $2{\times}10^{-6}$ & $0.2290 \pm 0.0003$ \\
 & AFHMC  & 10 & 1800 & $2{\times}10^{-5}$ & $0.2290 \pm 0.0004$ \\
 & AFHMC  & 50 & 1800 & $5{\times}10^{-6}$ & $0.2376 \pm 0.0008$ \\
 & AFHMC  & 100 & 1800 & $2{\times}10^{-6}$ & $0.2290 \pm 0.0004$ \\
\midrule
NLL & FA-LD  & 1 & 3600 & $1{\times}10^{-4}$ & $5445.18 \pm 136.26$ \\
 & FA-HMC & 10 & 3600 & $2{\times}10^{-5}$ & $4876.18 \pm 122.35$ \\
 & FA-HMC & 50 & 3600 & $5{\times}10^{-6}$ & $4544.43 \pm 34.72$ \\
 & FA-HMC & 100 & 3600 & $2{\times}10^{-6}$ & $4541.74 \pm 4.99$ \\
 & AFHMC  & 10 & 1800 & $2{\times}10^{-5}$ & $4532.89 \pm 7.66$ \\
 & AFHMC  & 50 & 1800 & $5{\times}10^{-6}$ & $4680.34 \pm 15.89$ \\
 & AFHMC  & 100 & 1800 & $2{\times}10^{-6}$ & $4531.58 \pm 7.61$ \\
\bottomrule
\end{tabular}}
\end{table}

\begin{table}[t]
\centering
\caption{\revise{Test accuracy on heterogeneous FMNIST at an equal number of server exchanges (AFHMC at round $x/2$, FA-LD and FA-HMC at round $x$). Entries are means over five random seeds; stepsizes as in Table~\ref{tab:fmnist-final-se}.}}
\label{tab:fmnist-trajectory}
\revise{\begin{tabular}{lccccccc}
\toprule
Method & $K$ & $\eta$ & $x=2000$ & $2400$ & $2800$ & $3200$ & $3600$ \\
\midrule
FA-LD  & 1 & $1{\times}10^{-4}$ & 0.8089 & 0.8085 & 0.8080 & 0.8076 & 0.8075 \\
FA-HMC & 10 & $2{\times}10^{-5}$ & 0.8270 & 0.8272 & 0.8273 & 0.8277 & 0.8274 \\
FA-HMC & 50 & $5{\times}10^{-6}$ & 0.8362 & 0.8369 & 0.8376 & 0.8381 & 0.8384 \\
FA-HMC & 100 & $2{\times}10^{-6}$ & 0.8369 & 0.8381 & 0.8391 & 0.8401 & 0.8408 \\
AFHMC  & 10 & $2{\times}10^{-5}$ & 0.8389 & 0.8403 & 0.8413 & 0.8419 & 0.8422 \\
AFHMC  & 50 & $5{\times}10^{-6}$ & 0.8373 & 0.8386 & 0.8391 & 0.8399 & 0.8402 \\
AFHMC  & 100 & $2{\times}10^{-6}$ & 0.8392 & 0.8404 & 0.8410 & 0.8416 & 0.8422 \\
\bottomrule
\end{tabular}}
\end{table}

\end{document}